\documentclass[10pt,twocolumn,letterpaper]{article}

\usepackage[pagenumbers]{wacv} 

\usepackage{graphicx}
\usepackage{amsmath}
\usepackage{amssymb}
\usepackage{booktabs}
\usepackage{array}
\usepackage{caption}
\usepackage{subcaption}
\usepackage[symbol]{footmisc}

\usepackage{multirow}

\usepackage[pagebackref,breaklinks,colorlinks]{hyperref}

\usepackage[capitalize]{cleveref}
\crefname{section}{Sec.}{Secs.}
\Crefname{section}{Section}{Sections}
\Crefname{table}{Table}{Tables}
\crefname{table}{Tab.}{Tabs.}

\def\wacvPaperID{1757} 
\def\confName{WACV}
\def\confYear{2024}

\begin{document}

\title{Dual-Camera Joint Deblurring-Denoising}

\author{
    Shayan Shekarforoush\normalfont{\textsuperscript{1, 4}}\thanks{Work done during an internship at Samsung AI Center Toronto} \and
    Amanpreet Walia\normalfont{\textsuperscript{2}}  \and
    Marcus A. Brubaker\normalfont{\textsuperscript{2, 3, 4}} \and
    Konstantinos G. Derpanis\normalfont{\textsuperscript{2, 3, 4}} \and
    Alex Levinshtein\normalfont{\textsuperscript{2}} \and
    \normalfont{
    \textsuperscript{1}University of Toronto \quad \textsuperscript{2}Samsung AI Center Toronto \quad \textsuperscript{3}York University \quad \textsuperscript{4}Vector Institute}
}
\maketitle

\begin{abstract}

Recent image enhancement methods have shown the advantages of using a pair of long and short-exposure images for low-light photography. These image modalities offer complementary strengths and weaknesses. The former yields an image that is clean but blurry due to camera or object motion, whereas the latter is sharp but noisy due to low photon count. Motivated by the fact that modern smartphones come equipped with multiple rear-facing camera sensors, we propose a novel dual-camera method for obtaining a high-quality image. Our method uses a synchronized burst of short exposure images captured by one camera and a long exposure image simultaneously captured by another. Having a synchronized short exposure burst alongside the long exposure image enables us to (i) obtain better denoising by using a burst instead of a single image, (ii) recover motion from the burst and use it for motion-aware deblurring of the long exposure image, and (iii) fuse the two results to further enhance quality. Our method is able to achieve state-of-the-art results on synthetic dual-camera images from the GoPro dataset with five times fewer training parameters compared to the next best method. We also show that our method qualitatively outperforms competing approaches on real synchronized dual-camera captures.
\href{https://shekshaa.github.io/Joint-Deblurring-Denoising/}{Project Webpage.}
\vspace{-0.4cm}
\end{abstract}

\section{Introduction}
\begin{figure}[t!]
\centering
\includegraphics[width=0.47\textwidth]{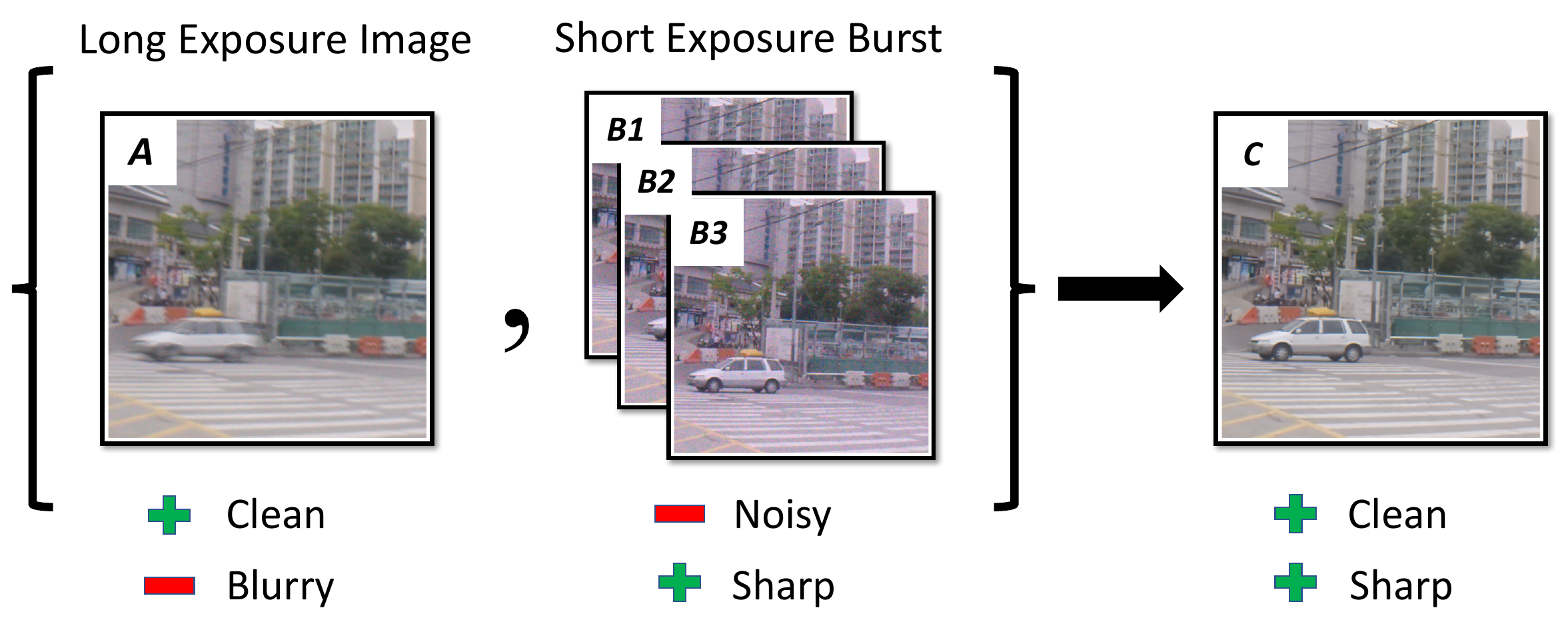}
\caption{Given a long exposure blurry image (A), synchronized with a burst of short exposure images (B1,B2,B3), we approach the joint deblurring-denoising task, producing a high quality clean and sharp image (C).
}
\vspace{-0.6cm}
\label{fig:short-long}
\end{figure}

Capturing high quality images under challenging conditions of low-light and dynamic scenes can be daunting, especially for smartphone cameras that have limited physical sensor size.
In such circumstances, the exposure time becomes a prominent factor that affects the quality of the final image.
Increasing the exposure time allows more light to reach the sensor, yielding an image with a lower level of noise and rich colors. 
However, long exposure causes objects to appear unpleasantly blurry, due to either camera or scene motion.
While lowering exposure time helps capture sharp details at high sensitivity settings (ISO), the resulting image becomes susceptible to noise and color distortion.
This trade-off is illustrated in Fig.\ \ref{fig:short-long}.

Given a noisy or blurry image, one can recover an enhanced version using denoising or deblurring methods.
These restoration tasks are long-standing problems in image processing that
are often addressed independently using either traditional methods \cite{buades2005non, dabov2007image}, or recent learning-based ones \cite{zhang2017beyond, nah2017deep, gao2019dynamic} that owe their success to training convolutional neural networks on a large amount of clean/corrupted pairs of data.
To make use of complementary information from both types of images, some recent work aim for the joint task of deblurring-denoising when corresponding short and long exposure images are available as input \cite{mustaniemi2018lsd,chang2021low,zhao2022d2hnet,zhang2022self}.
These multi-image methods outperform single image baselines.

Many modern smartphones have multiple rear-facing cameras and acquisition of synchronized images is becoming an emerging capability \cite{lai2022face} (multi-camera Android API).
In this paper, we employ a two-camera imaging system for joint deblurring and denoising.
Similar to earlier classical work \cite{tai2008image, nayar2004motion}, we acquire a burst of short exposure images, rather than a single image, with a synchronized long exposure image.
The benefit is two-fold. First, under the assumption of relative rigidity between the cameras, temporal synchronization of captures enables non-blind deblurring of the long exposure image based on the motion information in the burst \cite{tai2008image, nayar2004motion}.
Some previous work have addressed the joint deblurring-denoising task in either synchronized \cite{lai2022face} or unsynchronized \cite{mustaniemi2018lsd, chang2021low, zhao2022d2hnet} settings; however, cannot infer the motion from the single short exposure image.
Some others \cite{tico2021deep}, despite capturing multiple images with different exposure times, are limited to a single camera without synchronization.
Second, compared to the two-frame joint deblurring-denoising approaches \cite{mustaniemi2018lsd,chang2021low,zhao2022d2hnet,zhang2022self, lai2022face}, complimentary information across multiple independent noisy images in the burst can be leveraged to more accurately estimate the underlying signal \cite{mildenhall2018burst, xia2020basis}.

To this end, we first adapt a motion-aware deblurring network \cite{zhang2021exposure}, equipped with deformable convolutions, to exploit externally provided optical flow; so the network deblurs a given long exposure image in a non-blind fashion.
We compute the optical flow between frames in the burst using a pre-trained network.
In case of spatial misalignment of cameras in a real dual-camera system, we align images before being passed to the method.
In addition to the new flow-guided deblurring architecture, we employ a lightweight version of burst processing \cite{bhat2021deep} to denoise short exposure images into a single clean version.
We then fuse the resulting denoised and deblurred intermediate features, and a final clean and sharp image is reconstructed. 
We train the entire pipeline, consisting of deblurring, denoising and fusion, in an end-to-end fashion and evaluate the performance on synthetic data constructed by repurposing the GoPro dataset \cite{nah2017deep}, as well as data obtained using a real hardware-synchronized dual-camera system.
Our contributions are summarized as follows:
\begin{itemize}
     \item A novel flow-guided deblurring network, that uses the motion in the short exposure burst to remove non-uniform motion blur in the long exposure image.
     \item A joint architecture for burst denoising and deblurring, combining complementary information from the short exposure burst and long exposure image for the final image restoration.
     
\end{itemize}
Our approach to joint deblurring-denoising achieves state-of-the art results on challenging synthetic data and performs competitively with others on real data.
We will release our code upon publication.

\section{Related Work}
\label{sec:related}

\textbf{Image Deblurring.}
Image blur can arise from camera shake, object motion, focus, or a combination of these factors \cite{fergus2006removing,ito2014blurburst,abuolaim2020defocus,sun2015learning}. 
Some early non-blind deconvolution methods, like Lucy-Richardson \cite{lucy1974iterative,richardson1972bayesian} and Wiener deconvolution \cite{brown1992introduction}, aim to restore the image assuming a known blur kernel. Other traditional deblurring methods aim to recover the sharp latent image (and blur kernel) from a blurry image using optimization techniques. 
Most techniques make assumption about camera motion \cite{gupta2010single,whyte2012non,zheng2013forward} or use image priors \cite{fergus2006removing,jia2007single,chen2019blind,shan2008high,cho2009fast,xu2010two,pan2016l_0} for regularization.
Although these techniques can perform well under controlled settings justifying the chosen prior, their application to real-world images is limited. 
With recent advances of deep neural networks, learning-based methods \cite{levin2009understanding,sun2012super,kohler2012recording,lai2016comparative,nah2017deep,shen2019human,rim2020real,jiang2020learning,su2017deep,nah2019ntire,hradivs2015convolutional,shen2018deep} are able to achieve state-of-the-art results even in challenging scenarios. CNN-based methods can recover the deblurred image (and blur kernels) from a blurry image either in an end-to-end fashion \cite{gong2017motion,noroozi2017motion,nah2017deep,kupyn2019deblurgan,purohit2020region,zhang2021exposure,yuan2020efficient} or using deconvolution from predicted blur kernels \cite{xu2017motion,sun2015learning}. 
Some methods predict the deblurring result directly from the blurry image in a blind manner e.g.,\cite{noroozi2017motion}, \cite{nah2017deep}, \cite{kupyn2019deblurgan}. 
Other works \cite{gong2017motion}, \cite{purohit2020region}, \cite{yuan2020efficient}, \cite{zhang2021exposure} decompose the deblurring problem into predicting motion from the blurry image and using it for non-blind image deblurring. We follow the latter approach, but use flow predicted from a synchronized short exposure burst, resulting in a more reliable motion estimate.


 
\textbf{Image Denoising.}
Classical denoising methods aim to reduce noise in an image by applying filtering either in spatial \cite{tomasi1998bilateral,buades2005non,dabov2007image,takeda2007kernel} or frequency \cite{portilla2003image} domains. 
Similar to image deblurring, deep learning-based denoising methods \cite{anwar2019real,guo2019toward,yu2019deep,zhang2017beyond,zhang2018ffdnet} have been shown to outperform traditional methods. 
More complex noise synthesis models \cite{guo2019toward,wei2020physics,monakhova2022dancing} can further boost denoising performance.

Low-light imaging has especially benefited from burst processing methods. A burst of short exposure images allows the capture of more light by integrating information across the burst, while overcoming the problem of motion blur and non-uniform dynamic ranges pronounced in a long exposure image. 
Some approaches do not compute the burst motion explicitly or represent it implicitly using spatially varying kernels \cite{tassano2020fastdvdnet,kokkinos2019iterative,mildenhall2018burst,xia2020basis}, relying on end-to-end deep learning to integrate information across the burst. Others make use of explicit motion estimation \cite{godard2018deep, lecouat2021lucas, bhat2021deep, bhat2021deepiccv} using Lucas-Kanade \cite{lucas1981iterative} or deep optical flow methods. With a moderate amount of noise, when motion can still be estimated reliably, the latter methods achieve state-of-the-art performance \cite{luo2022bsrt}. In our work, we make use of a burst denoising method with an explicit motion estimate \cite{bhat2021deep}. In addition to obtaining good denoising performance, we reuse the computed motion for flow-guided long exposure image deblurring and fuse the results.

\textbf{Image Restoration using Short and Long Exposure Images.}
To exploit complementary information available in short and long exposure images, some methods combine the two.
Early dual-camera approaches \cite{nayar2004motion,tai2008image} propose a hybrid imaging system that uses predicted motion from one camera to deblur the long exposure image from a second camera. 
Others classical methods like \cite{yuan2007image} and \cite{zhang2013multi} uses degraded image pairs (noisy or blurry) to recover sharp latent image. 
Recently, deep learning-based methods have been used to combine a single long and short exposure image for image restoration \cite{mustaniemi2018lsd,chang2021low,zhao2022d2hnet,lai2022face}. 
LSD2 \cite{mustaniemi2018lsd} processes concatenated short and long exposure images with a simple U-Net style architecture. 
LSF \cite{chang2021low} proposes a more advanced joint data synthesis pipeline that improves results on real data. 
D2HNet \cite{zhao2022d2hnet} proposes a two-phase architecture trained on synthetically generated long/short exposure pairs to first deblur a long-exposure image and then enhance the result based on the guidance from a single noisy short-exposure image.
Lai et al.\ \cite{lai2022face} 
engineered a dual-camera system on Google's Pixel to synchronously capture frames for face deblurring.
They train deep CNNs to align and fuse a blurry but low noise raw image from the main camera with a sharp but noisy raw image from the secondary camera, to generate a final clean and sharp image of the face region.

Similar to the traditional methods \cite{nayar2004motion,tai2008image} we use a dual-camera system to enhance image quality. 
However, we employ modern deep learning approaches for motion-guided deblurring, and fuse the deblurring and burst denoising results to further boost image quality.
\section{Problem Definition}
\label{sec:problem}
\begin{figure}
    \centering
    \includegraphics[width=0.45\textwidth]{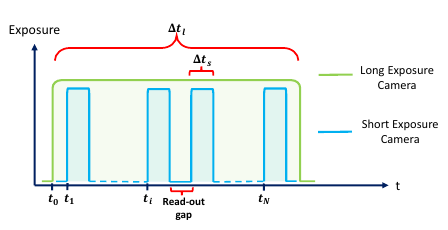}
    \caption{Visualization of temporal synchronization of two captures in an ideal dual-camera setting. The short and long exposure times are denoted as $\Delta t_s$ and $\Delta t_l$, respectively. Read-out gaps cause missing information in the burst.}
    \vspace{-0.5cm}
    \label{fig:exposure}
\end{figure}

In the joint denoising-deblurring task, the input consists of ${H \times W}$ resolution sRGB images including a burst of short exposure images, $\{S_i\}_{i=1}^{N}$, and a single blurry image, $L$. 
In a dual-camera setting, recording short and long exposure images can occur at different temporal orders.
Here, we assume that one camera captures the long exposure image while a second camera simultaneously acquires a burst.
In other words, as illustrated in Fig.\ \ref{fig:exposure}, the entire burst capture occurs within the time span of the long-exposure one.
Due to physical limitations of the camera, 
there are read-out gaps between frames of the burst causing missing information, as opposed to the long exposure camera that keeps recording the light during the entire capture.




Assuming that cameras are temporally synchronized, the clean raw measurements of long and short exposure images, respectively denoted by $R$ and $R_i$, are formalized as
\begin{equation}
  R = \int_{t=t_0}^{t_0 + \Delta t_l} I(t)dt, \quad R_i = \int_{t=t_i} ^ {t_i + \Delta t_s} I(t) dt,
  \label{eq:raw}
\end{equation}
where $I(t)$ is the incoming light to the sensor at time $t$. 
The capture of the $i$-th image in the burst starts at $t_i$ and ends after $\Delta t_s$. Meanwhile, the long exposure spans an interval of $\Delta t_l$ starting from $t_0$.
Due to sensor and photon noise, recorded raw images are noisy. These are then processed by the camera's Image Signal Processing pipeline (ISP) to produce visually appealing images, $\{S_i\}_{i=1}^{N}$ and $L$.

The final goal is to adopt pairs of burst and long exposure images $(\{S_i\}_{i=1}^{N}, L)$, to restore a single clean and sharp image, aligned with a reference image in the burst. In particular, we define the ground-truth image as $G=\text{ISP}(R_m)$, where $m$ is the index of the middle image considered as reference.
We set the total number of images in the burst to be odd, namely $N=2k-1$, implying $m=k$.

\section{Methodology}
\label{sec:method}
Our architecture is shown in Fig.\ \ref{fig:methods}.
We first discuss our approach to each individual task, deblurring and denoising, and then describe a fusion process trained end-to-end that combines the results into a final sharp and clean output.
\begin{figure}[t]
    \centering
    \includegraphics[width=0.47\textwidth]{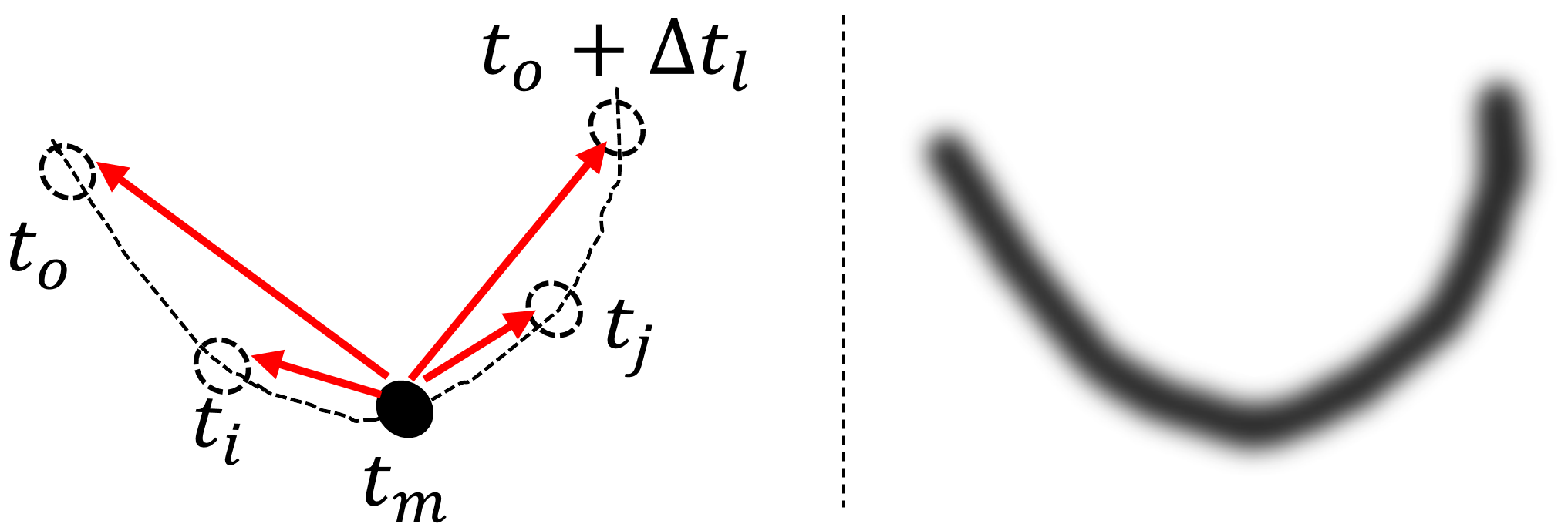}
    \caption{Deblurring given a motion trajectory. On the left, a single moving point from a latent sharp scene is observed at different positions along a motion trajectory (dotted line). In a long exposure image (right), this results in a blur streak aligned with the motion trajectory. Deblurring can be achieved by integrating the information along the motion trajectory.}
    \label{fig:blur_streak}
    \vspace{-0.45cm}
\end{figure}

\begin{figure*}
    \hspace{0.5cm}
    \centering
    \includegraphics[width=0.91\textwidth]{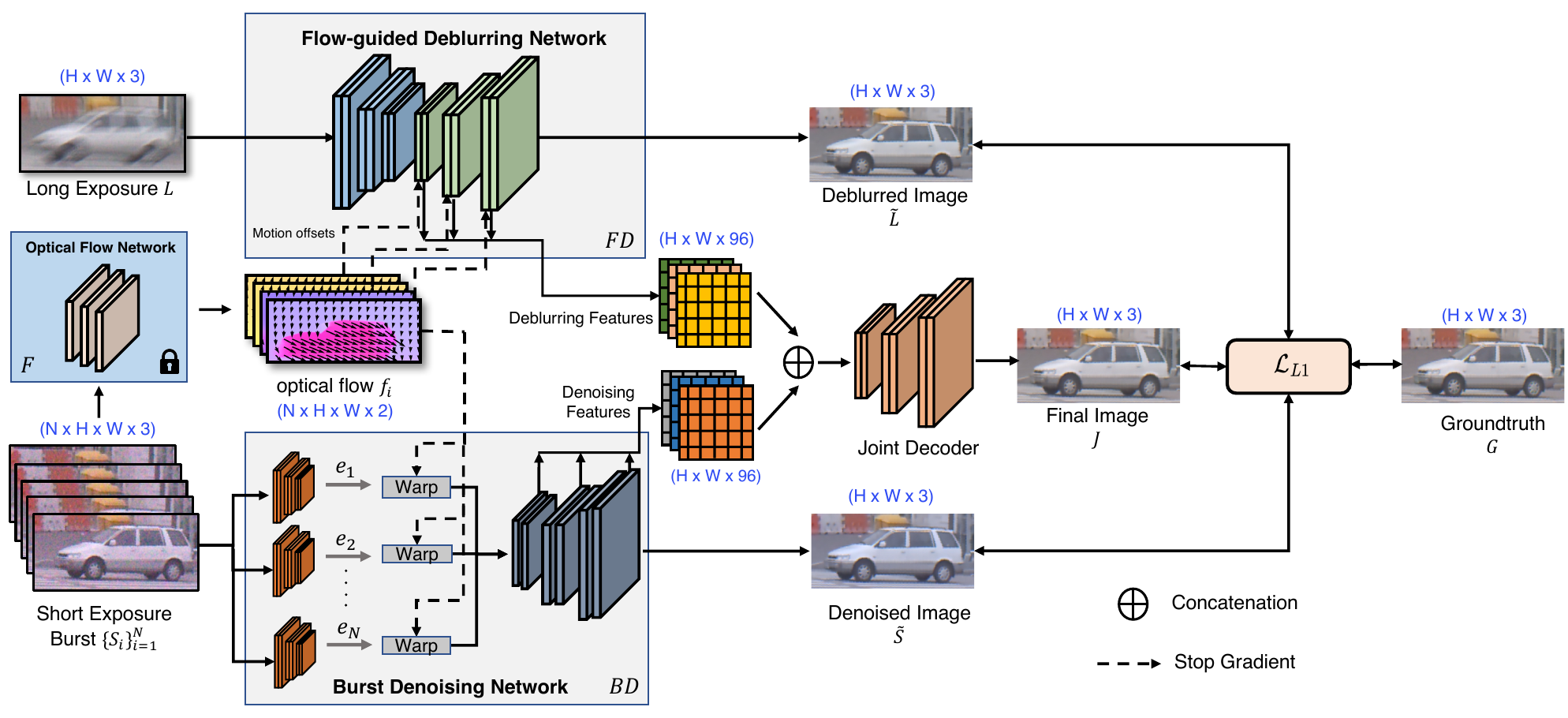}
    \caption{Overview of our architecture. Our workflow is composed of three parts. (i) A motion-aware deblurring module ($\text{FD}$), guided by freezed optical flows obtained from the burst, that removes blur from the given long exposure image. (ii) A denoiser ($\text{BD}$), conditioned on the same optical flows, which merges the burst into a clean image (iii) A fusion module that combines feature representations from deblurring and denoising and decodes the result to a final deblurred-denoised image.}
    \vspace{-0.5cm}
    \label{fig:methods}
\end{figure*}
\subsection{Flow-guided Deblurring}
\label{sec:deblur}
In real dynamic scenes, motion blur is highly non-uniform and thus removing it requires the use of spatially varying kernels.
Following Motion-ETR \cite{zhang2021exposure}, we adopt a fully-convolutional deblurring network, enabled to incorporate motion information that adaptively modulates the shape of convolution kernels. 
Using flexible kernels is motivated by a fundamental result from previous works \cite{zhang2018dynamic, purohit2020region}: the deconvolution of a blurry image requires filters with similar direction/shape as the blur kernel.
Accordingly, to determine the shape of filters, we employ the exposure trajectory model \cite{zhang2021exposure}, an extension to the blur kernel with time dependency.
The exposure trajectory characterizes how a point is displaced from the reference frame at different timesteps.
Fig.~\ref{fig:blur_streak} illustrates this for a single moving point in a scene. 
During capture, a point is observed at varying locations relative to the reference position at time $t_m$. 
A long exposure time yields a blurry streak aligned with the motion trajectory. 
Given the motion trajectory for a point, deblurring is achieved by integrating the information along.

To recover the trajectory for a given point, we need to obtain spatial offsets expressing shifts from the middle time step to other time steps.
Thanks to the temporal synchronization between the burst and long exposure image, we leverage the motion observed in the burst to obtain discrete samples of the motion trajectories (red arrows in Fig.~\ref{fig:blur_streak}).
While Motion-ETR \cite{zhang2021exposure} trains a separate network to predict relative offsets conditioned on the input blurry image, we use the optical flow between frames in the noisy burst.
In particular, for a pixel, $\mathbf{p}$, in the reference frame, $S_{m}$, the motion to an arbitrary frame, $S_i$, can be represented as
\begin{equation}
    S_{m}(\mathbf{p}) = S_{i}(\mathbf{p} + \Delta\mathbf{p_i}),
\end{equation}
where $\Delta\mathbf{p_i}$ denotes the flow vector for the point $\mathbf{p}$.
As illustrated in Fig. \ref{fig:blur_streak}, these time dependent flow vectors across the burst, $\Delta\mathbf{p}=\{\Delta\mathbf{p_i}\}_{i=1}^{N}$, describe the trajectory of the corresponding pixel.
We argue that our pre-trained flows are more accurate offsets than those provided by Motion-ETR.
First, optical flow networks are typically trained and fine-tuned over a large amount of data, likely to achieve better generalization to new scenes.
Second, our offsets are computed from sharp frames in the burst while those in Motion-ETR are estimated from the blurry image.
We support our argument using examples in the results (\ref{sec:ablation}).

Once flow vectors are obtained, we linearly interpolate them into a trajectory with $K^2$ points (currently $K=3$) and reshape the result into $K \times K$ deformable convolution kernels \cite{dai2017deformable}. 
Thus, kernels have spatially varying support across the image domain. 
We use the same backbone architecture as Motion-ETR \cite{zhang2021exposure}, which is based on DMPHN \cite{zhang2019deep}. 
In this hierarchical network, the last convolution at each level is deformable.
Given the input blurry image, $L \in \mathbb{R}^{H \times W \times 3}$, we formulate the deblurring operation as:
\begin{equation}
    \label{eq:deblur}
    \tilde{L} = FD(L, \{\Delta\mathbf{p_i}\}_{i=1}^{N}; \theta_{FD}), \quad \Delta\mathbf{p_i} = F(S_i, S_m),
\end{equation}
where $FD$ denotes the flow-guided deblurring parameterized by $\theta_{FD}$ and the optical flow network, $F$, returns flow vectors, $\Delta\mathbf{p_i} \in \mathbb{R}^{H \times W \times 2}$, computed between short exposure images and the reference frame. We choose PWC-Net \cite{sun2018pwc} as the flow estimator because of its high accuracy and speed. 
Empirically, we observe that this network is robust to the noise in the input burst and provides sufficiently accurate flows to be used in the downstream deblurring task.

\subsection{Burst Denoising}
\label{sec:burst-denoising}
For burst denoising, we adopt DBSR~\cite{bhat2021deep} due to its strong perfomance and use of optical flow. The latter allows the same optical flow to be used for both flow-guided deblurring and burst denoising.
Since, DBSR was originally designed for RAW burst super-resolution, we adapt it to the sRGB burst denoising task.
Our burst denoiser is composed of four modules described in turn.

\textbf{Encoder.} 
A shared encoder, $Enc$, is separately applied to each image, $S_i$, in the burst, obtaining individual feature representations, $e_i = Enc(S_i)$. 
We reduce the final feature dimensions from $D\hspace{-2pt}=\hspace{-2pt}512$ to $D\hspace{-2pt}=\hspace{-2pt}96$ to accommodate higher resolution training images ($256\hspace{-2pt}\times\hspace{-1pt}256$ rather than $96 \hspace{-2pt} \times \hspace{-2pt} 96$). 

\textbf{Alignment.} 
Due to camera and scene motion, images in the burst are spatially misaligned. 
To effectively fuse the burst features, $e_i$, we warp them to the reference frame, $S_{m}$ using the flow $\Delta \mathbf{p_i}$ in Eq.\ \ref{eq:deblur}. We obtain aligned features, $\tilde{e}_i = \phi(e_i, \Delta\mathbf{p_i})$,
where $\phi$ is the backwarp operation with bilinear interpolation.

\textbf{Fusion.} Once aligned, the features are combined across the burst to generate a final fused feature representation, $e \in \mathbb{R}^{H \times W \times D}$. 
We use the attention-based mechanism \cite{bhat2021deep} to adaptively extract information from images while allowing an arbitrary number of images as input.
Specifically, a weight predictor conditioned on warped features and flow vectors, is learned to return unnormalized log attention weights, $\tilde{w}_i \in \mathbb{R}^{H \times W \times D}$ for each warped encoding, $\tilde{e}_i$. 
The fused feature map is obtained using a weighted sum
\begin{equation}
    e = \sum_{i=1}^{N} \frac{exp(\tilde{w}_i)}{\sum_{j} exp(\tilde{w}_j)} \tilde{e}_i\ .
\end{equation}
See our supplement for details.

\textbf{Decoder.} 
The decoder reconstructs the final denoised version of the reference image from the the fused feature map, $\tilde{S}=Dec(e)$. 
We use a similar decoder architecture as DBSR \cite{bhat2021deep} but remove the upsampling layer from the decoder as we are not performing super-resolution.

The entire burst denoising can be summarized as,
\begin{equation}
    \tilde{S} = BD(\{S_i\}_{i=1}^{N}; \theta_{BD})\ ,
\end{equation}
where $BD$ is the burst denoiser network, $\tilde{S} \in \mathbb{R}^{H \times W \times 3}$ denotes the clean reconstructed image from the burst, and $\theta_{BD}$ contains all learnable parameters in all four modules.

\subsection{Joint Denoising-Deblurring}
\label{sec:joint}
We now define a learnable module that combines information from the flow-guided deblurring, $FD$, and the burst denoiser, $BD$, into a high quality image. 
This module receives intermediate feature representations from $FD$ and $BD$ modules, concatenates them into a single feature map and then decodes the result into the final clean and sharp image.
From the three-level hierarchical deblurring network, we choose the last layer
features 
of the decoders at all levels, yielding a feature map of total dimension $D_1 = 32 \times 3 = 96$. 
From the burst denoising branch, we select the fused feature map, $e$, of dimension $D_2 = 96$.
The concatenation is followed by a conv layer that merges the features into $D=96$ features. We then employ a decoder with the same architecture from Sec~\ref{sec:burst-denoising} but separate parameters to generate the final output for the joint task,
\begin{equation}
    J = Dec(\texttt{concat}(d, e); \theta_{J}),
\end{equation}
where $\texttt{concat}$ is the concatenation operator, $d \in \mathbb{R}^{H \times W \times 96}$ denotes the feature representation taken from the deblurring module, and $Dec$ is the decoder parameterized by $\theta_{J}$.

We train all modules in an end-to-end fashion by minimizing a multi-task weighted loss,
\begin{equation}
    \mathcal{L} = \mathcal{L}_1(J, G) + \lambda_{deblur}\mathcal{L}_1(\tilde{S}, G) + \lambda_{denoise}\mathcal{L}_1(\tilde{L}, G),
\end{equation}
where $\mathcal{L}_1$ is the average $L_1$ distance, and $G$ is the ground-truth image, as defined in Sec.\ \ref{sec:problem}.
While $\mathcal{L}_1(J, G)$ penalize the final joint output, $\mathcal{L}_1(\tilde{S}, G)$ and $\mathcal{L}_1(\tilde{L}, G)$ regularize intermediate deblurring and denoising outputs, respectively, with hyperparameters $\lambda_{deblur}$ and $\lambda_{denoise}$ determining their importance.
We study the impact of these additional loss terms in the experiments section.

\begin{figure}
  \includegraphics[width=0.48\textwidth]{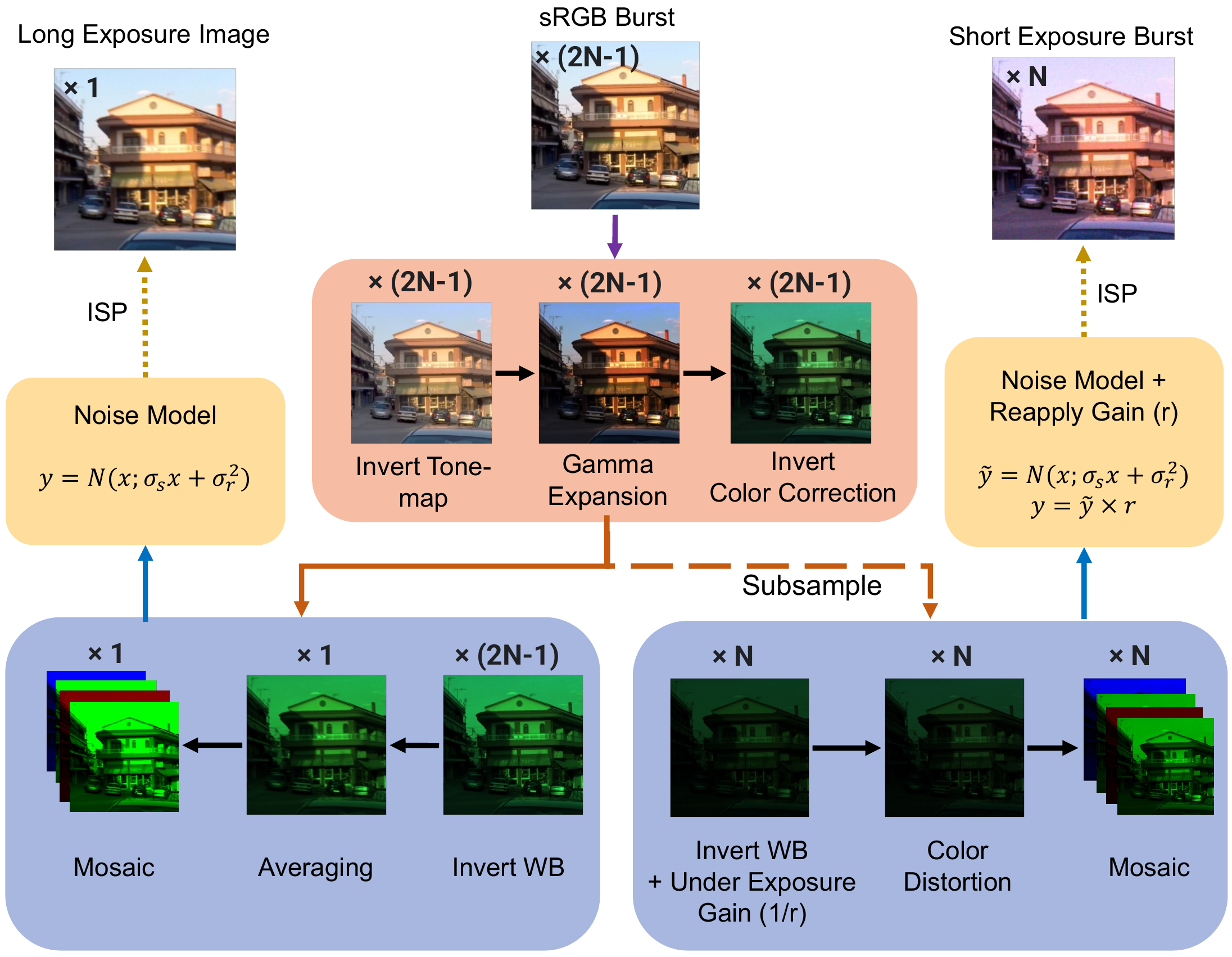}
  \caption{Data generation pipeline based on \cite{brooks2019unprocessing}. 
  Starting from a burst of $2N-1$ consecutive frames from GoPro, the tone mapping, gamma compression and color correction steps are inverted for all images.
  Afterwards, the pipeline branches into two - (i) The linearized intensities are averaged into a single raw image with realistic blur. (ii) The burst subsampled to simulate the read-out gap, followed by high level of noise addition and color distortion common in raw short exposure captures.
  Finally, the ISP is re-applied to obtain final processed long and short exposure images.}
  \vspace{-0.6cm}
  \label{fig:data-gen}
\end{figure}

\section{Datasets}
\textbf{Synthetic Data.}
To evaluate our method, we synthetically generate a dataset consistent with our problem definition (Sec.\ \ref{sec:problem}). 
We use the GoPro dataset \footnote{Images in Figs. 1, 4, 5, 7, 8, 10, 11, 12, 14 are available in \href{https://seungjunnah.github.io/Datasets/gopro}{GoPro dataset}~\cite{nah2017deep} under a \href{https://creativecommons.org/licenses/by/4.0/}{CC BY 4.0 License}.} which contains 33 high frame-rate $720 \times 1280$ resolution videos.
We follow the original data split of 22 videos for training and 11 videos for testing.
To generate pairs of realistic synchronized long and short exposure images as in Sec.\ \ref{sec:problem}, we use a procedure similar to \cite{chang2021low}.
To start, we obtain RAW images by \textit{unprocessing} \cite{brooks2019unprocessing} $2N-1$ consecutive frames.
On one hand, the resulting images are averaged to obtain the clean RAW long exposure image, while on the other, they are divided by the under-exposure factor $r$ to generate clean short exposure RAW burst.
To simulate read-out gaps in the short exposure burst, we skip every other frame, resulting in a $N$-frame burst.
In the original GoPro dataset, 
blurry frames were mostly generated by averaging $11$ consecutive frames. 
To remain consistent, while also simulating read-out gaps by dropping every other frame and having an odd number of burst frames, the maximum burst size is $N=5$.
The main results are presented with $N=5$. The impact of smaller burst sizes are also examined in the ablation study.
Subsequently, we add realistic degradations similar to \cite{chang2021low}. For the short exposure burst, we apply color distortion to simulate the commonly present purple tint similar to \cite{chang2021low, zhao2022d2hnet,mustaniemi2018lsd}. For both the long and short exposure images, we add heteroscedastic Gaussian noise. Finally, we process the images back into sRGB using an ISP. This results in triplets of long exposure image, short exposure burst, and ground-truth: $(L, \{S_i\}_{i=1}^{5}, G)$. Our full synthesis process is shown in Fig.~\ref{fig:data-gen}. More details can be found in the supplement.

\textbf{Real Data.}
\begin{figure}
\centering
  \includegraphics[width=0.4\textwidth]{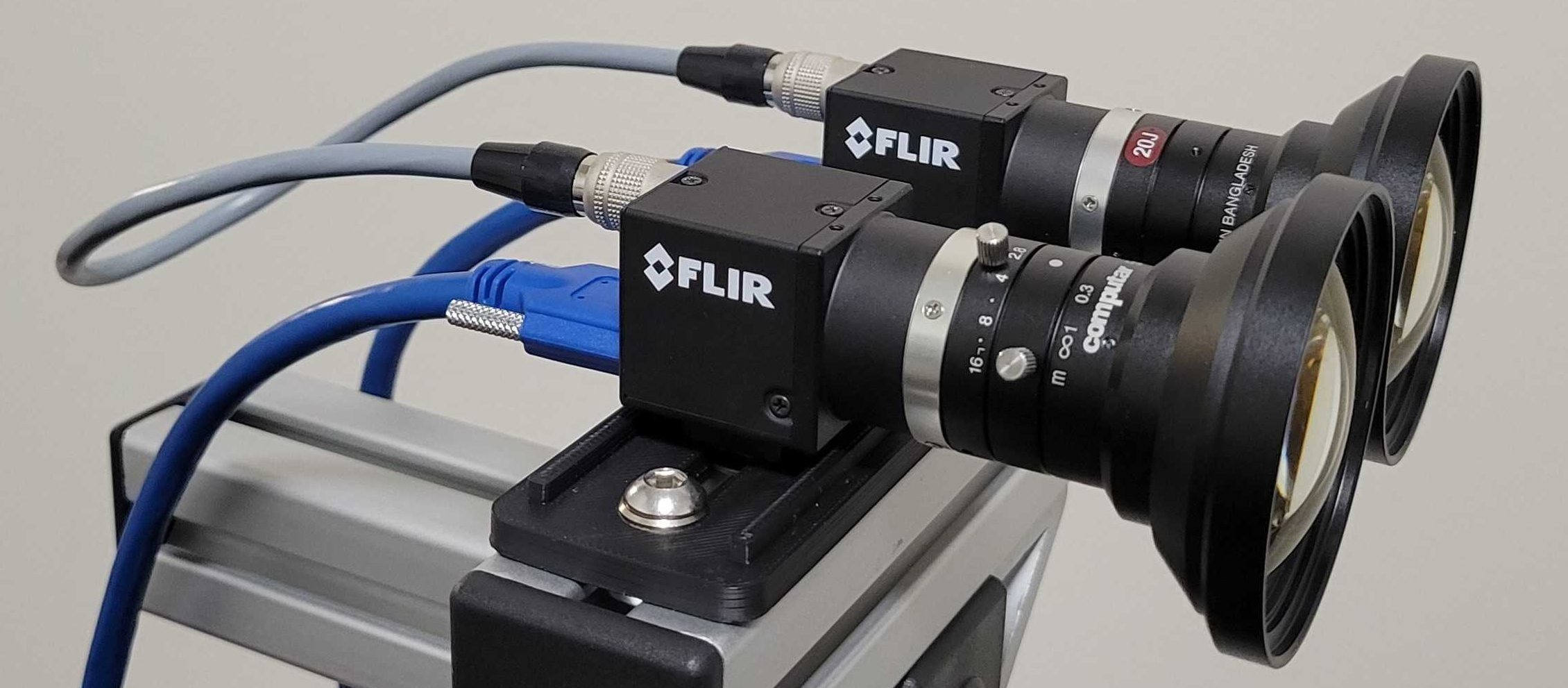}
  \caption{Synchronized dual-camera capture rig with FLIR BlackFly cameras, used to collected real data.}
  \label{fig:real-data}
  \vspace{-0.6cm}
\end{figure}
In addition to our synthetic data experiments, we capture several real synchronized long exposure images and short exposure bursts using the hardware synchronized FLIR camera rig setup shown in Fig.~\ref{fig:real-data}. 
Images are captured at $2048 \times 3072$ resolution in RAW format and processed with the ISP used for the synthetic data. 
Unlike the synthetic data, the two cameras are spatially misaligned. 
Thus prior to applying our method, the long exposure image is warped to the middle frame in the short exposure burst, using a RANSAC-based robust homography fitting.

\begin{figure*}
    \begin{subfigure}[b]{0.5\textwidth}
         \centering
         \includegraphics[width=\textwidth]{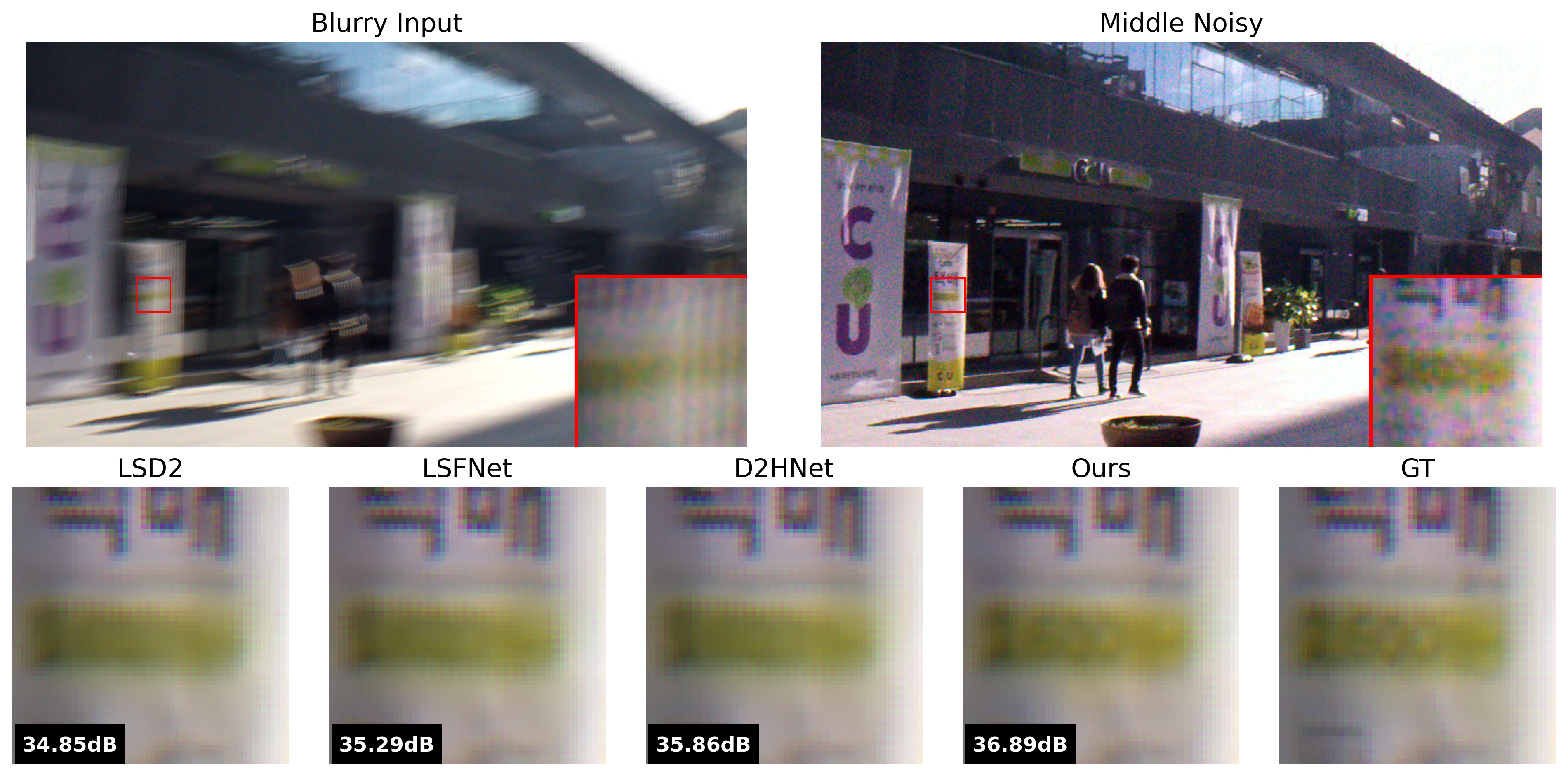}
     \end{subfigure}
    \begin{subfigure}[b]{0.5\textwidth}
         \centering
         \includegraphics[width=\textwidth]{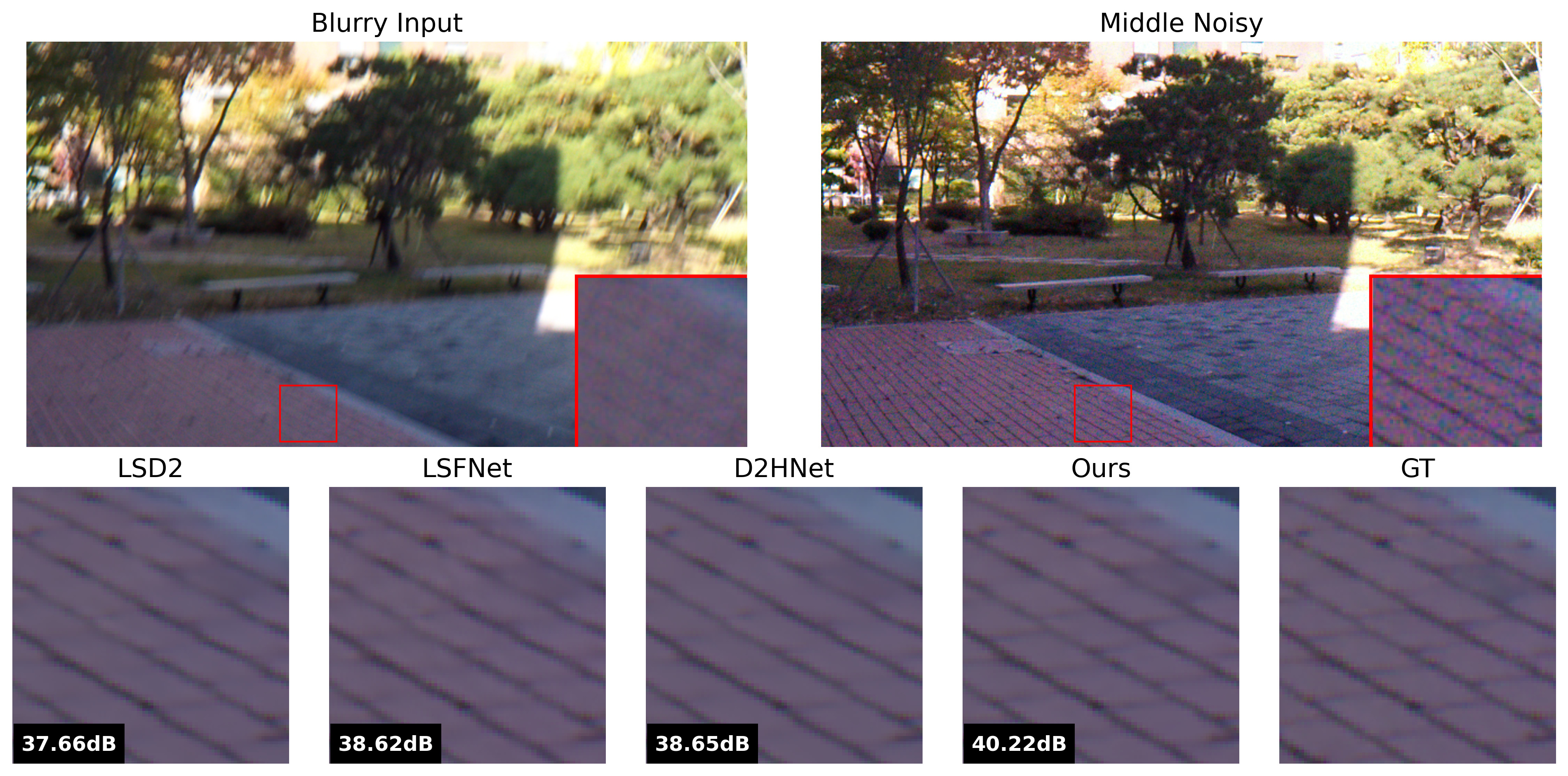}
     \end{subfigure}
     \caption{Qualitative comparison of our method with the state-of-the-art on two examples of test synthetic data.}
     \vspace{-0.45cm}
     \label{fig:vs_others}
\end{figure*}
\section{Experiments}

\subsection{Implementation Details.}
We train our method using an AdamW optimizer \cite{loshchilov2018decoupled} with a cosine annealing learning rate scheduler \cite{loshchilov2016sgdr} starting from $lr=0.0001$ to $lr=0.00001$ for a total of $200$ epochs.
The training data consists of $256 \times 256$ patches passed in mini-batches of size $B=16$ to the network. 
During training, we do not update the pre-trained optical flow network. 
We set $\lambda_{deblur}=\lambda_{denoise}=0.25$ unless otherwise specified.
Experiments are implemented in PyTorch and models are trained on four NVIDIA A40 GPUs for 15 hours.

\subsection{Joint Deblurring-Denoising}
To our knowledge, there is no previous work addressing the task of joint deblurring-denoising while using burst of short exposure images as input.
Nevertheless, a handful of methods \cite{mustaniemi2018lsd, zhao2022d2hnet, chang2021low, zhang2022self, lai2022face} approach a similar task where, rather than a burst, only a single short-exposure image is available.
Classical works \cite{nayar2004motion,tai2008image} are also close to our setting proposing a hybrid camera to correct the blur in a single long exposure image based on the motion blur estimated from multiple short exposure frames captured at the same time.
However, they use short exposure images solely for motion blur estimation, whereas we further process them into a denoised high quality image.
Through experiments on our synthetically generated dataset, we examine the image restoration quality of our proposed method and compare it with the SOTA: LSD2 \cite{mustaniemi2018lsd}, LSF \cite{chang2021low}, and D2HNet \cite{zhao2022d2hnet}.

\textbf{Experimental setting.} 
Our problem setting differs from that in each of the above work and we modify the corresponding training procedure or method to have a fair evaluation.
First, the baseline joint methods inherently accept only a single short exposure image in addition to the long exposure one.
Hence, for the baselines, we pair the middle frame of the burst with the blurry image, jointly passed as the input.
In contrast, our method leverages the entire burst. 
There are also baseline-specific differences.
For example, the input and output images for LSF are in RAW space. 
Thus, during training and inference, it requires an ISP to post-process the output into an sRGB image.
To maintain consistency in training and evaluation across all methods, we train the LSF network on our synthetic data with sRGB images as input and eliminate any post-processing.
LSF also uses a pixel shuffle layer to output an image at $\times 2$ resolution of the input. 
As our inputs and outputs match in size, we remove the pixel shuffle layer.
Since the D2HNet model is large, training it from scratch on our synthetic dataset results in overfitting. Instead, we fine-tuned two different versions of D2HNet on our synthetic dataset, both of which are pretrained on the large dataset from \cite{zhao2022d2hnet}. We fine-tune one model on patches (row 1 of Table \ref{tab:joint}) and another on full-sized images (row 2 of Table \ref{tab:joint}).

\textbf{Results.} 
After bringing all methods into a consistent setting, we evaluate their performance on our synthetic data. 
In addition to the two common metrics of PSNR and SSIM \cite{wang2004image}, we use LPIPS \cite{zhang2018unreasonable} as a perceptual metric to quantitatively assess performance.
For the joint task of deblurring-denoising, quantitative results on test data are reported in Table\ \ref{tab:joint}. 
LSD2 achieves the lowest performance, despite having a relatively large number of learnable parameters. 
In contrast, LSF use more complex layers such as deformable convolutions, but overall require less parameters. 
Our method significantly outperforms LSD2 and LSF in terms of PSNR. 
D2HNet fine-tuned on full-resolution synthetic data considerably outperforms that trained on patches of size $256 \times 256$.
Although D2HNet is pre-trained on more data and have significantly more parameters, our method achieves superior performance in terms of all metrics.
Qualitative examples are also demonstrated in Fig.\ \ref{fig:vs_others}.
In the first example, the simulated motion and noise in the input image inhibits perceiving the number written on the poster. 
Unlike other methods, our method is able to recover the number better.
This is also confirmed by the higher PSNR measured within the selected region.
In the other example, high frequency details such as cracks between tiles are the target; our method is able to recover all of them, unlike others.
Additional examples are provided in the supplement.

\begin{table}
\caption{Comparison of our method with Joint Deblurring-Denoising SOTA in terms of PSNR, SSIM and LPIPS metrics. Best results are shown in \label{tab:joint}
\textbf{bold}.}
\vspace{-5pt}
\centering
\resizebox{\columnwidth}{!}{%
\begin{tabular}{l|ccc|c}
\textbf{Method} & \textbf{PSNR}$\uparrow$ & \textbf{SSIM}$\uparrow$ & \textbf{LPIPS}$\downarrow$ & \textbf{\# params}     \\ \hline
\textsc{D2HNet}\footnotemark\cite{zhao2022d2hnet}          & 31.30         & 0.933         & 0.114          & \multirow{2}{*}{82.3M} \\
\textsc{D2HNet}\footnotemark\cite{zhao2022d2hnet}          & 37.87         & 0.987         & 0.026          &                        \\ \hline
\textsc{LSD2}\cite{mustaniemi2018lsd}            & 36.42         & 0.981         & 0.035          & 31M                    \\
\textsc{LSF}\cite{chang2021low}             & 37.22         & 0.984         & 0.033          & 10.9M                  \\ \hline
\textsc{Ours}
            & \textbf{38.25}         & \textbf{0.988}         & \textbf{0.025}          & 17.4M\\ 
\end{tabular}%
}
\vspace{-0.65cm}
\end{table}
\footnotetext[1]{Fine-tuned with same input patch size used in \cite{zhao2022d2hnet}}
\footnotetext[2]{Fine-tuned with full-sized images}
\begin{figure*}
    \begin{subfigure}[b]{0.49\textwidth}
         \centering
         \includegraphics[width=\textwidth]{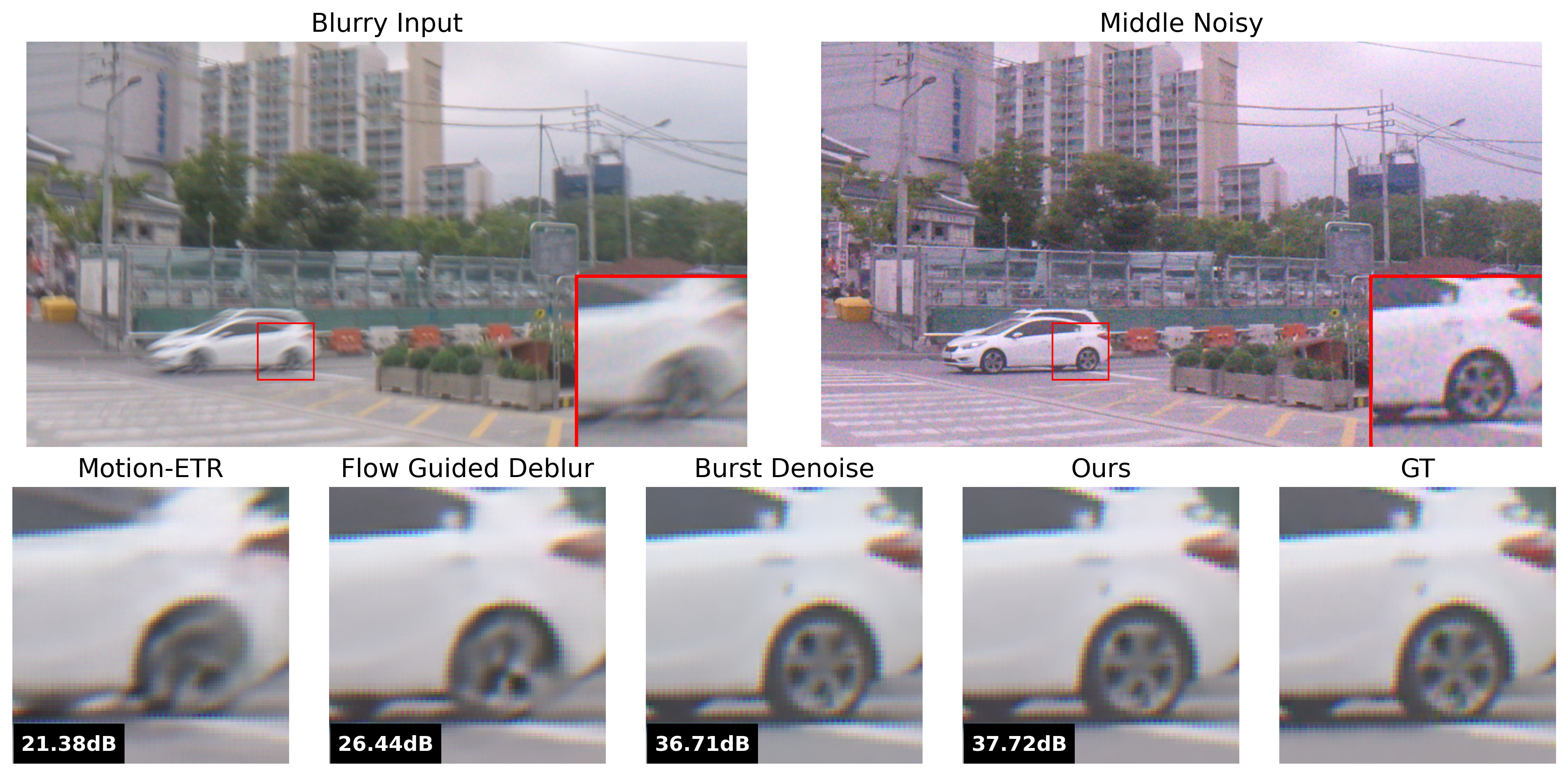}
     \end{subfigure}
    \begin{subfigure}[b]{0.49\textwidth}
         \centering
         \includegraphics[width=\textwidth]{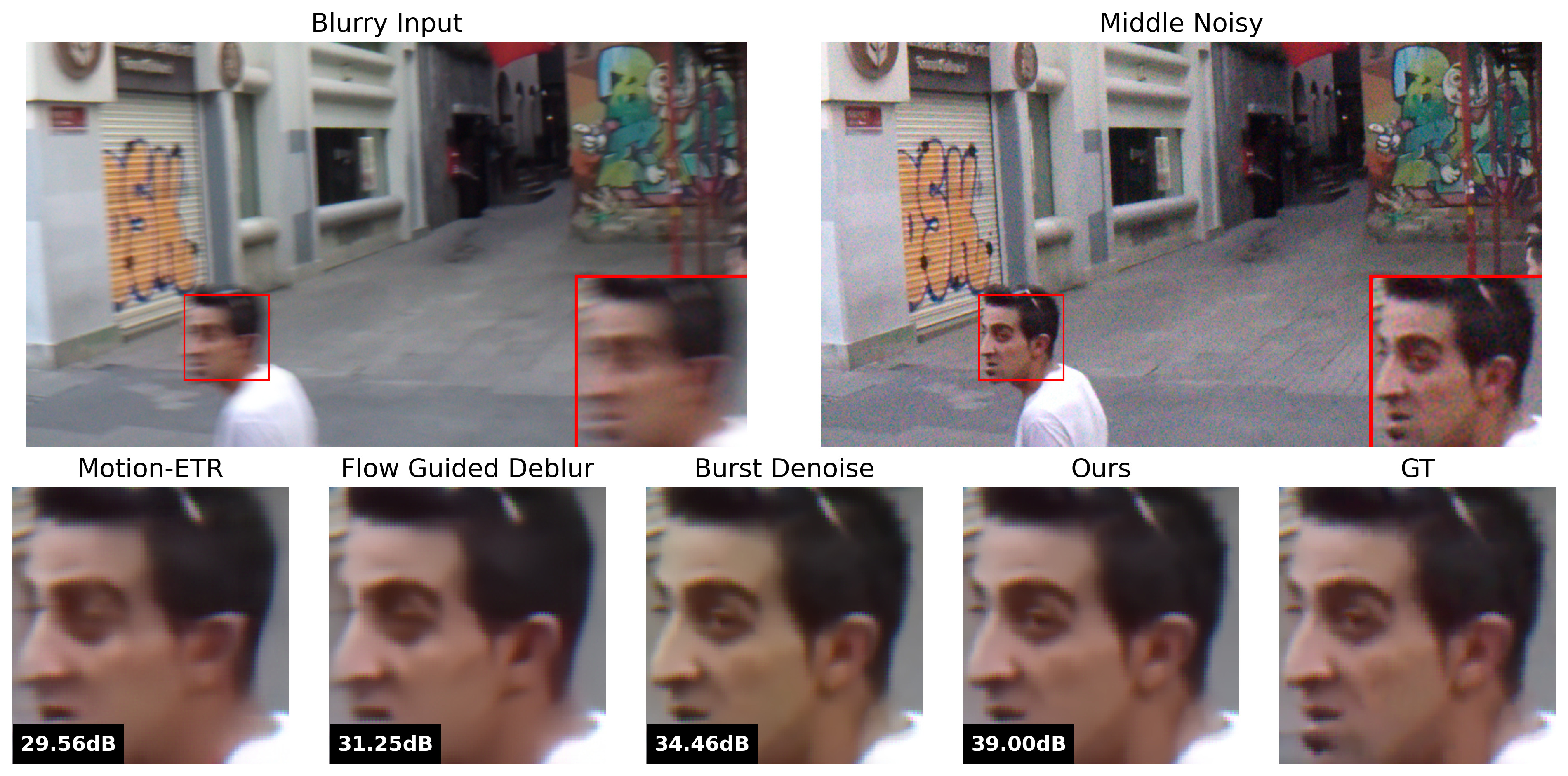}
     \end{subfigure}
     \caption{Qualitative comparison of our method in the ablation study on two example of test synthetic data.}
     \label{fig:ablation}
     \vspace{-0.2cm}
\end{figure*}
\begin{figure}
    \begin{subfigure}[b]{0.50\textwidth}
         \centering
         \includegraphics[width=\textwidth]{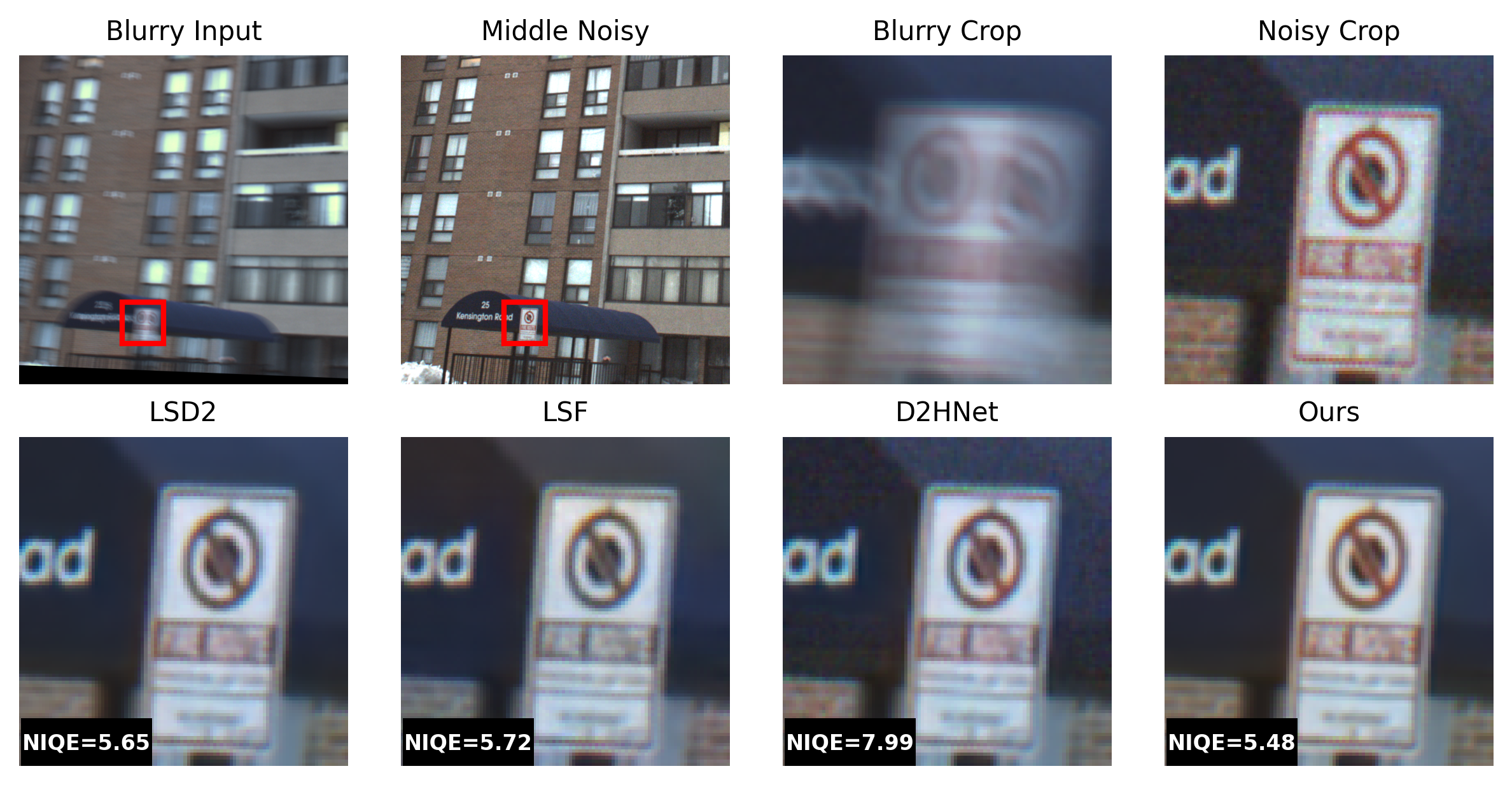}
     \end{subfigure}\\
    \begin{subfigure}[b]{0.50\textwidth}
         \centering
         \includegraphics[width=\textwidth]{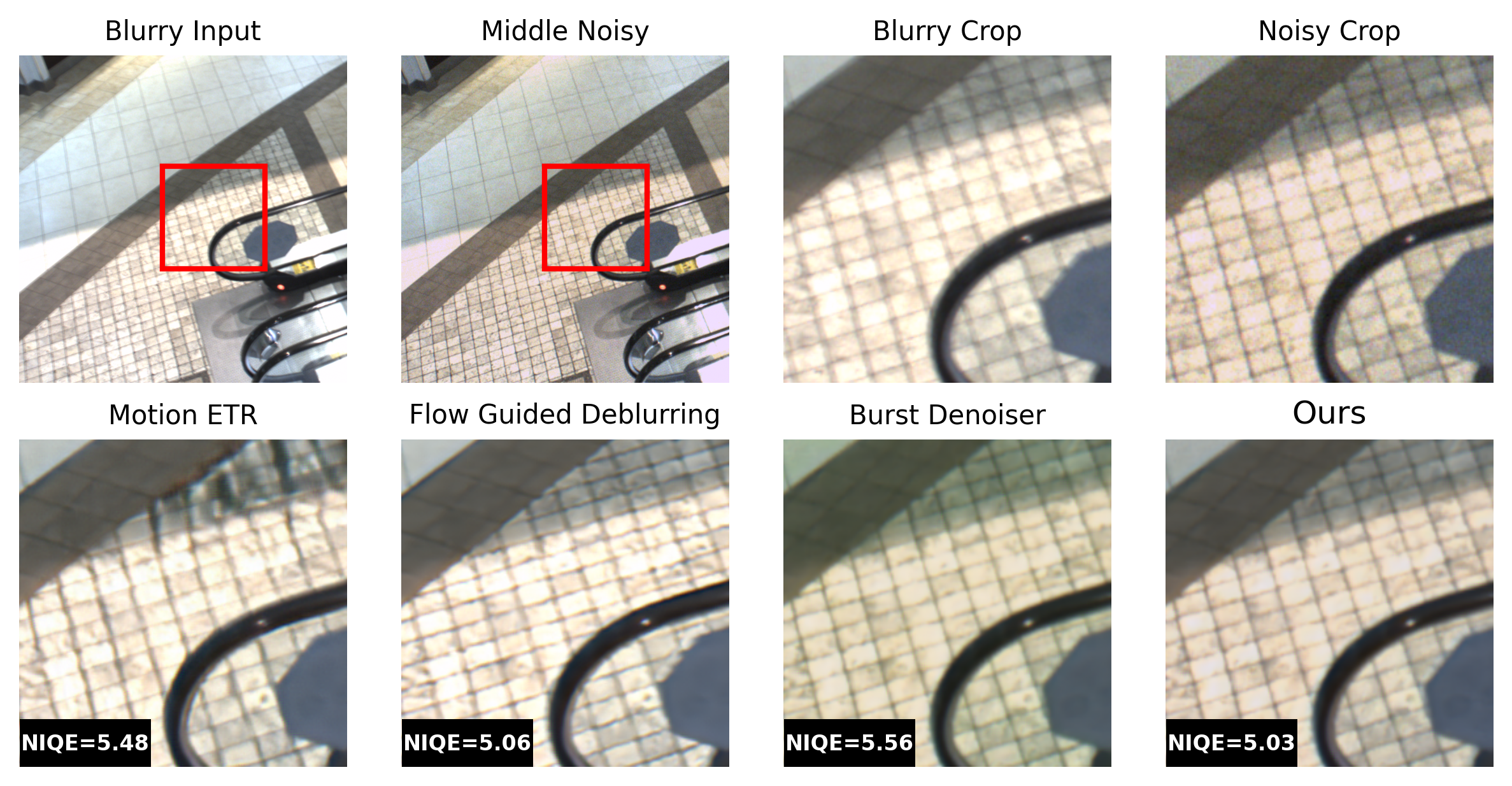}
     \end{subfigure}
     \caption{Qualitative comparison of different methods on two data examples captured by our real dual-camera system.}
     \label{fig:real_results}
\end{figure}
\begin{figure}
     \includegraphics[width=0.48\textwidth]{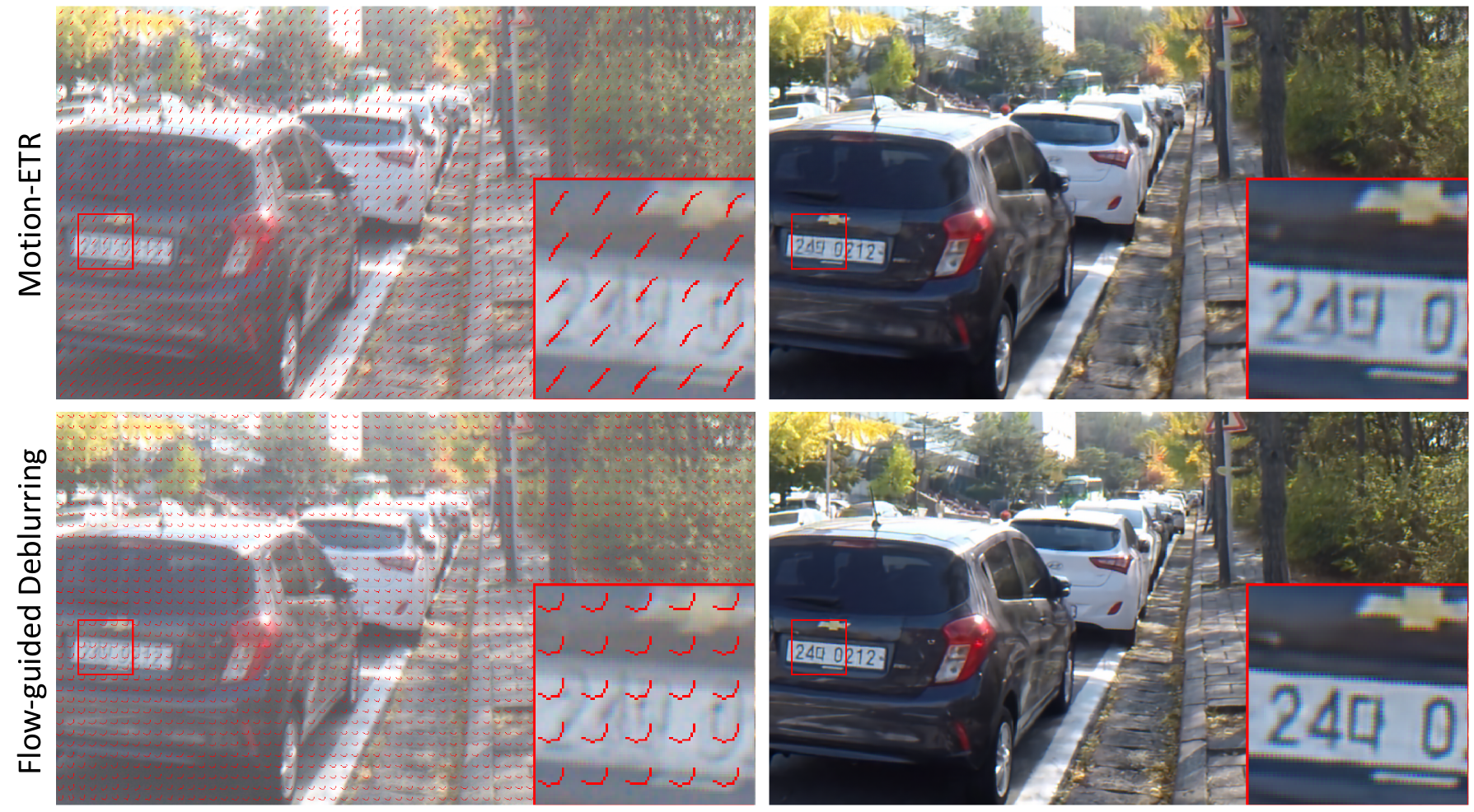}
     \caption{Visualization of trajectories provided to Motion-ETR \cite{zhang2021exposure} and ours flow-guided deblurring, overlaid on top of the input blurry image.
     The respective network outputs are also displayed.}
     \vspace{-0.55cm}
     \label{fig:flow_vis}
\end{figure}
\subsection{Ablation Study}
\label{sec:ablation}
\textbf{Auxiliary losses.}
We first explore the impact of intermediate loss terms $\mathcal{L}_1(\tilde{S}, G)$ and $\mathcal{L}_1(\tilde{L}, G)$.
We train our method with the multi-task loss using different combinations, $\lambda_{deblur}, \lambda_{denoise} \in \{0, 0.25\}$. 
When $\lambda=0$, the corresponding term is effectively eliminated.
The quantitative results in the Table \ref{tab:lambda_burst} indicates that the best PSNR is achieved by the setting $\lambda_{deb.} =\lambda_{den.}=0.25$, while differences in SSIM and LPIPS are marginal.
Also, the performance degrades with higher values of $\lambda=0.5$.

\textbf{Burst size.}
We also examine the effect of burst size $N$.
We evaluate our method in a new experiment where every other frame is kept from the burst of size $N\hspace{-2pt}=\hspace{-2pt}5$ yielding a new burst of size $N\hspace{-2pt}=\hspace{-2pt}3$.
We also examine the trivial case $N\hspace{-2pt}=\hspace{-2pt}1$ by keeping only the middle frame. To train our method for $N\hspace{-2pt}=\hspace{-2pt}1$, we drop the flow guidance for deblurring, and feature warping and attention in the burst-processing.
These quantitative results in Table \ref{tab:lambda_burst} show that PSNR significantly drops when reducing the burst size to $N=3\hspace{-2pt}$.
For $N\hspace{-2pt}=\hspace{-2pt}1$ even larger degradation is observed.
Note that our joint model with $N\hspace{-2pt}=\hspace{-2pt}3$ still outperforms prior work of LSD2, LSF and D2HNet, while being more efficient than the $N\hspace{-2pt}=\hspace{-2pt}5$ version as requiring less input images.
\begin{table}
\caption{Effect of regularization terms and burst size.}
\label{tab:lambda_burst}
\vspace{-7pt}
\footnotesize
\centering
\resizebox{\columnwidth}{!}{%
\begin{tabular}{c|cc|ccc}
N & $\lambda_{deb.}$ & $\lambda_{den.}$ & \textbf{PSNR}$\uparrow$ & \textbf{SSIM}$\uparrow$ & \textbf{LPIPS}$\downarrow$ \\ \hline
5 & 0   & 0   & 38.14  & 0.9882  & 0.0254 \\
5 & 0   & 0.25 &   38.19    &   0.9882     &    0.0252      \\
5 & 0.25 & 0   &    38.17   &    0.9883    &       0.0251   \\
5 & 0.25 & 0.25   &  \textbf{38.25}     &    \textbf{0.9883}    &      \textbf{0.0250}    \\
5 & 0.5 & 0.5   &  38.18    &   0.9882   &   0.0254  \\
\hline
3 & 0.25 & 0.25   &   37.94 &   0.987   & 0.028 \\
1 & 0.25 & 0.25   &     37.28   &  0.984    & 0.033 \\
\end{tabular}%
}
\vspace{-0.74cm}
\end{table}

\textbf{Single-Task baselines.}
Next, we explore how individual flow-guided deblurring and burst denoising modules perform compared to the joint model.
We also compare our new deblurring sub-module with Motion-ETR to validate the improvement achieved by replacing learned offsets with pre-trained optical flow.
We train flow-guided deblurring and burst denoiser networks separately and then evaluate them on the synthetic data.
For Motion-ETR, we follow the original training protocol \cite{zhang2021exposure}.
Quantitative results are summarized in Table\ \ref{tab:baselines}, showing that the joint model noticeably outperforms the individual modules. 
This verifies that our simple fusion process is able to take advantage of both worlds and output a more accurate image in the joint restoration task.
Motion blur is often more difficult to tackle than noise, as reflected in the superiority of burst denoising to deblurring.
Qualitative examples are also shown in Fig.\ \ref{fig:ablation}. 
The burst denoiser outputs a sharp image while flow-guided deblurring is slightly better at maintaining colors close to the ground-truth.
When integrated in our joint model, the final restoration does not suffer from color distortion and remains sharp.
Finally, flow-guided deblurring significantly outperforms the Motion-ETR in all provided metrics.
In Fig.\ \ref{fig:flow_vis}, we use spatial offsets to visualize the trajectory that a sparse grid of pixels follow during the exposure time. 
Comparing these trajectories shows that the pre-trained flow is capable of recovering highly non-linear motion, while the quadratic trajectory learned by Motion-ETR is not expressive enough, leading to a deblurred result of lower quality.  


\begin{table}
  \caption{Comparison of our method with baselines approaching individual tasks in terms of PSNR, SSIM and LPIPS metrics}
     \label{tab:baselines}
     \vspace{-5pt}
\centering
\resizebox{\columnwidth}{!}{
\begin{tabular}{l|c|ccc}
\textbf{Method}                  & \textbf{Task}            & \textbf{PSNR}$\uparrow$  & \textbf{SSIM}$\uparrow$  & \textbf{LPIPS}$\downarrow$ \\ \hline
\textsc{Motion-ETR}\cite{zhang2021exposure}            & Deblurring      & 30.17 & 0.936 & 0.102 \\ \hline
\multirow{3}{*}{\begin{tabular}[c]{@{}c@{}}\textsc{Ours}\\ (flow-guided)\end{tabular}} & Deblurring      & 32.23 & 0.958 & 0.078 \\ \cline{2-5}
                      & Burst-Denoising & 35.92 & 0.986 & 0.032 \\ \cline{2-5}
                      & Joint           & \textbf{38.25} & \textbf{0.988} & \textbf{0.025} \\ 
\end{tabular}
}
\vspace{-20pt}
\end{table}
\subsection{Real Data}
Finally, we evaluate all methods on real data captured by the dual-camera system. 
The models are trained on the synthetic data that does not perfectly match the real one in every aspect. 
Despite this gap, as shown in Fig.~\ref{fig:real_results}, our approach produces sharper and cleaner results while other methods suffer from noise, incorrect colors or severe artifacts.
Compared to our flow-guided deblurring, Motion-ETR \cite{zhang2021exposure} performs worse in removing the blur. 
Note that the individual burst denoiser outputs incorrect colors, whereas our method is able to compensate for it by using deblurring branch.
We also report
Natural Image Quality Evaluator (NIQE) \cite{mittal2012making}, a non-reference-based metric commonly used for image quality assessment when the ground-truth is not available. 
A lower value indicates better quality. 
We achieve the lowest value in all the provided examples.
Please see the supplement for further examples.



\vspace{-0.15cm}
\section{Conclusion}
In this paper, we presented a new method for joint image deblurring-denoising using a burst of short exposure images synchronized with a long exposure image, captured using a dual-camera rig.
Our flow-guided deblurring method uses the motion information from the burst for non-blind long exposure image deblurring.
We reuse the same flow estimate in a burst denoiser to output a clean image, whose features are then fused with that of deblurring into an improved final clean and sharp image.
We provide a proof-of-concept
that integrating flow-guided deblurring with burst denoising can achieve notable improvements upon previous joint task works on both synthetic data and real data. 
Deployment on phones with tailored engineering is part of future work.

\section*{Acknowledgements}
This work was done during an internship at the Samsung AI Center in Toronto, Canada. SS's internship
was funded by Mitacs Accelerate.
We would like to thank Abhijith Punnappurath and Michael Brown for the discussion on data generation.

{\small
\bibliographystyle{ieee_fullname}
\bibliography{egbib}
}
\clearpage
\appendix
\begin{center}
\Large
{\bf
Supplementary Material \\ [0.5em]
}
\end{center}
%
\section*{Outline}
Here, we outline our supplemental material. In Sec.\ \ref{sec:data}, we elaborate on our synthetic data generation pipeline. 
We discuss details of the noise model and blurry data generation that help simulating long and short exposure images. 
In Sec.\ \ref{sec:results}, we provide additional qualitative examples of our method and comparisons with our baselines, including single-task ones, on both synthetic and real data. 
In the \href{https://shekshaa.github.io/Joint-Deblurring-Denoising/}{Project Webpage}, we include an interactive demo.
Finally, Sec.\ \ref{sec:arch} outlines details of our network architecture.

\begin{figure*}
    \centering
  \includegraphics[width=0.7\textwidth]{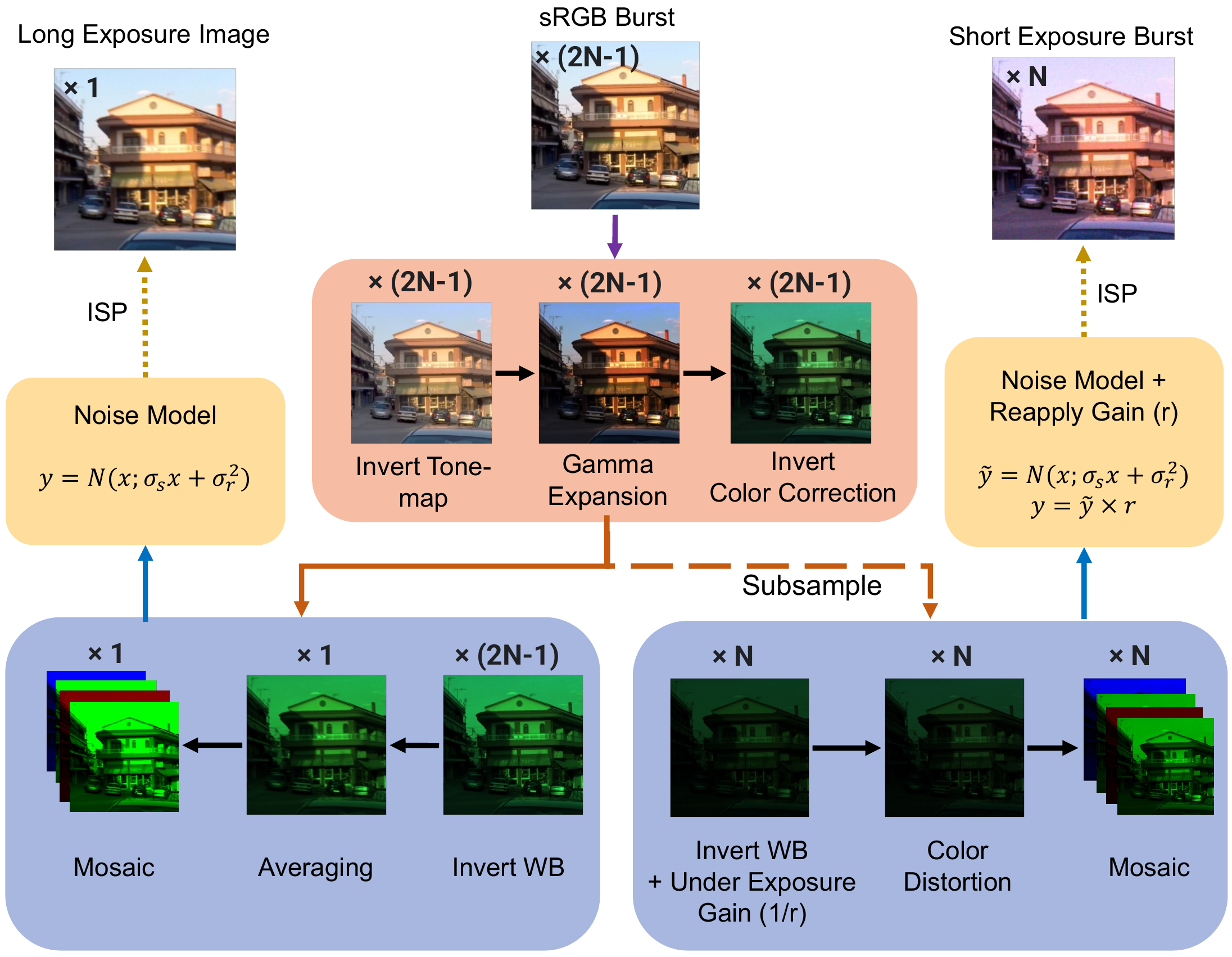}
  \caption{Data generation pipeline inspired by \cite{brooks2019unprocessing}. 
  Starting from a burst of $2N-1$ consecutive frames from GoPro, the tone mapping, gamma compression and color correction steps are inverted for all images.
  Afterwards, the pipeline branches into two. (i) The first branch averages the linear intensities to generate a single raw image with realistic blur. (ii) A separate branch sub-samples the burst to simulate the read-out gap, plus adding high level of noise and color distortion common in raw short exposure captures.
  Finally, the ISP is re-applied to obtain final processed long and short exposure images.}
  \label{fig:data-gen-supp}
  \vspace{-0.5cm}
\end{figure*}

\section{Synthetic Data Generation}
\label{sec:data}
We synthetically generate triplets of short exposure burst, long exposure image, and corresponding ground-truth image, $(\{S_i\}_{i=1}^{N}, L, G)$, complying with our problem setting.
Our data originates from the GoPro dataset \cite{nah2017deep} which is not readily amenable to our problem setting. 
First of all, in our realistic setting, we demand both long and short exposure images be noisy whereas the provided frames are clean. 
Second, recall the following equation from our problem definition, 
\begin{equation}
  R = \int_{t=t_0}^{t_0 + \Delta t_l} I(t)dt, \quad R_i = \int_{t=t_i} ^ {t_i + \Delta t_s} I(t) dt.
\end{equation}
Considering small read-out gaps, the raw long exposure image is approximated as the sum across raw short exposure images. However, in the current GoPro dataset, blurry images are constructed in an unrealistic fashion, by simply averaging sharp frames in the sRGB space.

We aim to simulate the noise in both short and long exposure images, $(\{S_i\}_{i=1}^{N}, L)$, and the blur in the long exposure image, $L$, by modeling them in the raw space.
To do so, first we obtain the corresponding raw measurements for the provided sRGB images.
Following the steps in previous work \cite{brooks2019unprocessing}, we \textit{unprocess} images, which essentially converts images from the sRGB space back to the raw space. 
Our unprocessing scheme first inverts the tone-map and applies gamma expansion, $\lambda=2.2$, to compute a linear RGB image. 
Cameras also transform RGB color measurements to sRGB values using $3 \times 3$ color correction matrices (CCMs), which is needed to be inverted after gamma expansion.
The above steps, as illustrated on Fig.\ \ref{fig:data-gen-supp}, are shared between short and long exposure image generation. Hereafter, we require separate treatment to construct the short exposure burst and the long exposure image.

For the burst, we first simulate read-out gaps by taking every other frame from the sequence. Then the white balance (WB) is inverted by dividing red and blue channels by random gains drawn from intervals $[1.9, 2.4]$ and $[1.5, 1.9]$, respectively. 
Similar to previous work \cite{brooks2019unprocessing}, for saturated pixels, instead of naive division by gain parameters, we apply a customized transformation that keeps their intensity high. In addition to WB inversion, images in the burst are divided by a fixed exposure ratio, ${r} = 10$, to simulate the underexposure effect with the same special treatment for saturated pixels as in the WB inversion.
Moreover, it is common to see color distortion in short exposure images with high ISO, appearing as purplish artifacts \cite{chang2021low, zhao2022d2hnet}. 
We simulate such effects as done in previous work \cite{chang2021low} by multiplying the red and blue channels by random values in $[1.0, 1.1]$. 
All these burst-specific steps are depicted in the right branch of Fig.\ \ref{fig:data-gen-supp}. 
On the other hand, the branch for the long exposure image receives all frames in the sequence and inverts WB gains. 
Next, the average is computed across the WB inverted burst to obtain a single blurry raw image. 
Note that the averaging in raw space is closer to realistic captures than previous methods \cite{zhao2022d2hnet, mustaniemi2018lsd, chang2021low, zhang2022self} which either simulate blur using gyroscope data or perform the averaging in sRGB space.
In both branches, assuming the Bayer pattern of RGGB, a simple mosaic step is ultimately applied to images producing the clean Bayer raw counterparts.

Next, we simulate sensor noise to be added to raw short and long exposure images. 
The noise in raw data stems from two main sources: (i) shot noise that accounts for randomness in photon arrivals, and (ii) read noise due to readout effects \cite{healey1994radiometric}.
Shot noise is a Poisson random variable whose mean and variance depend on the intensity level, while read noise is a zero-mean Gaussian random variable with fixed variance. 
As done previously \cite{brooks2019unprocessing}, the two noise sources can be modeled as a single heteroscedastic Gaussian whose variance depends on the intensity level. 
Namely, for each clean pixel value, $x_p$, the noise pixel, $y_p$, is
\begin{equation}
    y_p = x_p + n_p, \quad n_p \sim \mathcal{N}(0, \sigma_s x_p + \sigma_r^2),
    \label{eq:noise}
\end{equation}
where $N(m,s^2)$ is a normal distribution with mean $m$ and variance $s^2$. Analog and digital gains, denoted as $g_a$ and $g_d$, respectively, control short and read noise parameters, $\sigma_s$ and $\sigma_r^2$, as follows \cite{healey1994radiometric, brooks2019unprocessing}:
\begin{equation}
    \sigma_s = g_a g_d, \quad \sigma_r^2 = g_a^2 g_d^2.
\end{equation}
We use a similar noise model as prior work \cite{brooks2019unprocessing} which randomly generates $\log(\sigma_s)$ and $\log(\sigma_r^2)$ in the log scale:
\begin{align}
    \log(\sigma_s) &\sim U(0.000125, 0.0002), \\
    \log(\sigma_r^2) | \log(\sigma_s) &\sim \mathcal{N}(\mu = 2.18 \log (\sigma_s) + 1.2, \sigma = 0.26),
\end{align}
where $U(a,b)$ denotes a uniform distribution ranging from $a$ to $b$.
The above is based on the noise calibration done by previous work \cite{brooks2019unprocessing}. 
They sample the shot noise parameter, $\log(\sigma_s)$, from a higher range, causing too much noise in our data. Thus, we adjust the range to accommodate our own data.
The same noise model is used in both, except for the burst branch to have the aforementioned exposure time ratio factor, $r$, applied on top.
This multiplicative scalar, $r$, can be interpreted as the ratio of ISO between two cameras.
Multiplying by $r$ ensures that the resulting noisy short exposure images have the same intensity level as as the corresponding noisy long exposure image.
This is equivalent to using a noise model with scaled digital gain $\tilde{g}_d = g_d \times r$ for the short exposure burst.
In fact, given a recorded intensity $x_p/r$ from a short exposure image,
\begin{equation}
    y_p = r(x_p / r + n_p) = x_p + r \times n_p = x_p + \tilde{n}_p,
\end{equation}
where,
\begin{align}
    n_p &\sim \mathcal{N}(0, \sigma_s x_p/r + \sigma_r^2), \\
    \tilde{n}_p &\sim \mathcal{N}(0, \tilde{\sigma}_s x_p + \tilde{\sigma}_r^2), \\
    \tilde{\sigma}_s &= r \times \sigma_s, \quad \tilde{\sigma}_r^2 = r^2 \times \sigma_r^2.
\end{align}

Finally, once noisy raw images are computed, we reapply the ISP to obtain the long and short exposure images in sRGB space. 
Our ISP applies demosaicing, white balance, color correction, gamma compression, and tone mapping, in turn.
We use simple bilinear interpolation for demosaicing \cite{brooks2019unprocessing}. 
For the white-balance, the corresponding red and blue gains are re-applied.
Color correction matrix (CCM) maps white-balanced images from camera RGB space to sRGB space.
Lastly, the gamma compression, $\lambda=2.2$, and the same tone-mapping as previous work \cite{brooks2019unprocessing} is used to obtain the final images.

\section{Additional Qualitative Results}
\label{sec:results}
Further qualitative examples on both real and synthetic data, comparing competitive methods with ours, are provided in this document.
We also encourage the reader to look at the \href{https://shekshaa.github.io/Joint-Deblurring-Denoising/}{interactive webpage}.

\begin{figure*}
    \begin{subfigure}[b]{\textwidth}
         \centering
         \includegraphics[width=\textwidth]{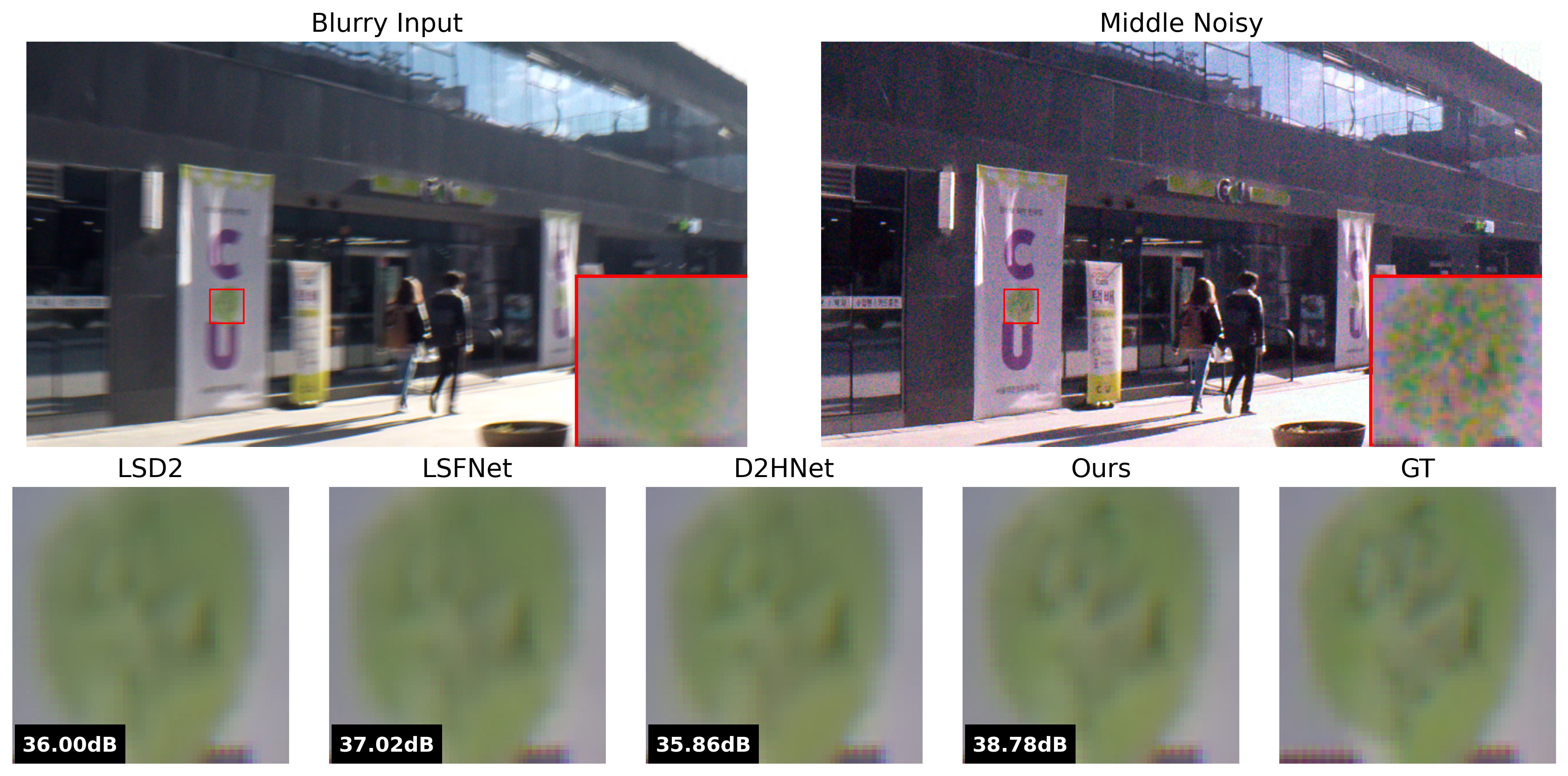}
     \end{subfigure}
    \begin{subfigure}[b]{\textwidth}
         \centering
         \includegraphics[width=\textwidth]{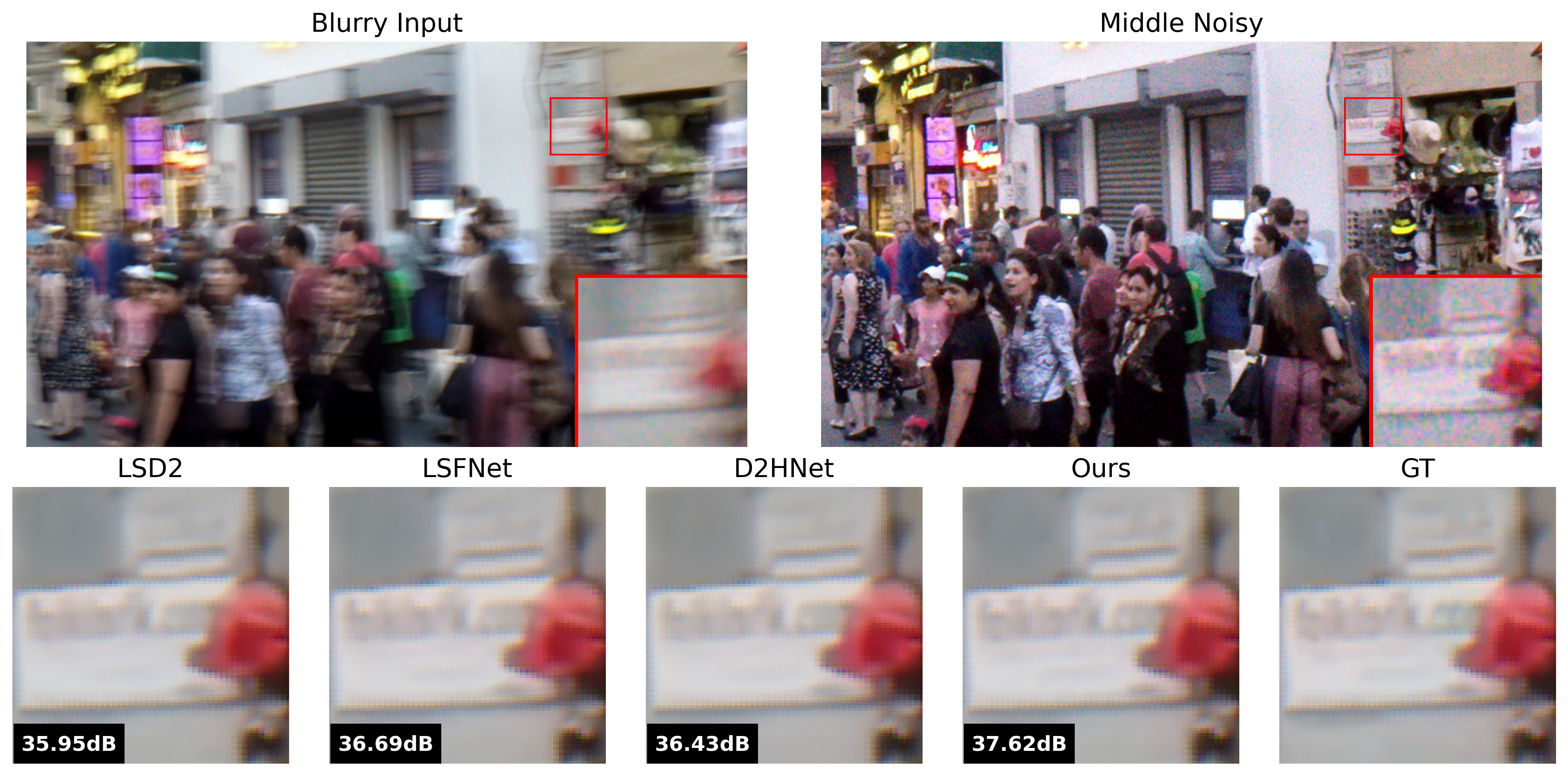}
     \end{subfigure}
     \caption{Qualitative comparison of our method with the state of the art on synthetic data examples.}
     \vspace{-0.15cm}
     \label{fig:supp_syn_joint}
\end{figure*}

\begin{figure*}
    \begin{subfigure}[b]{\textwidth}
         \centering
         \includegraphics[width=\textwidth]{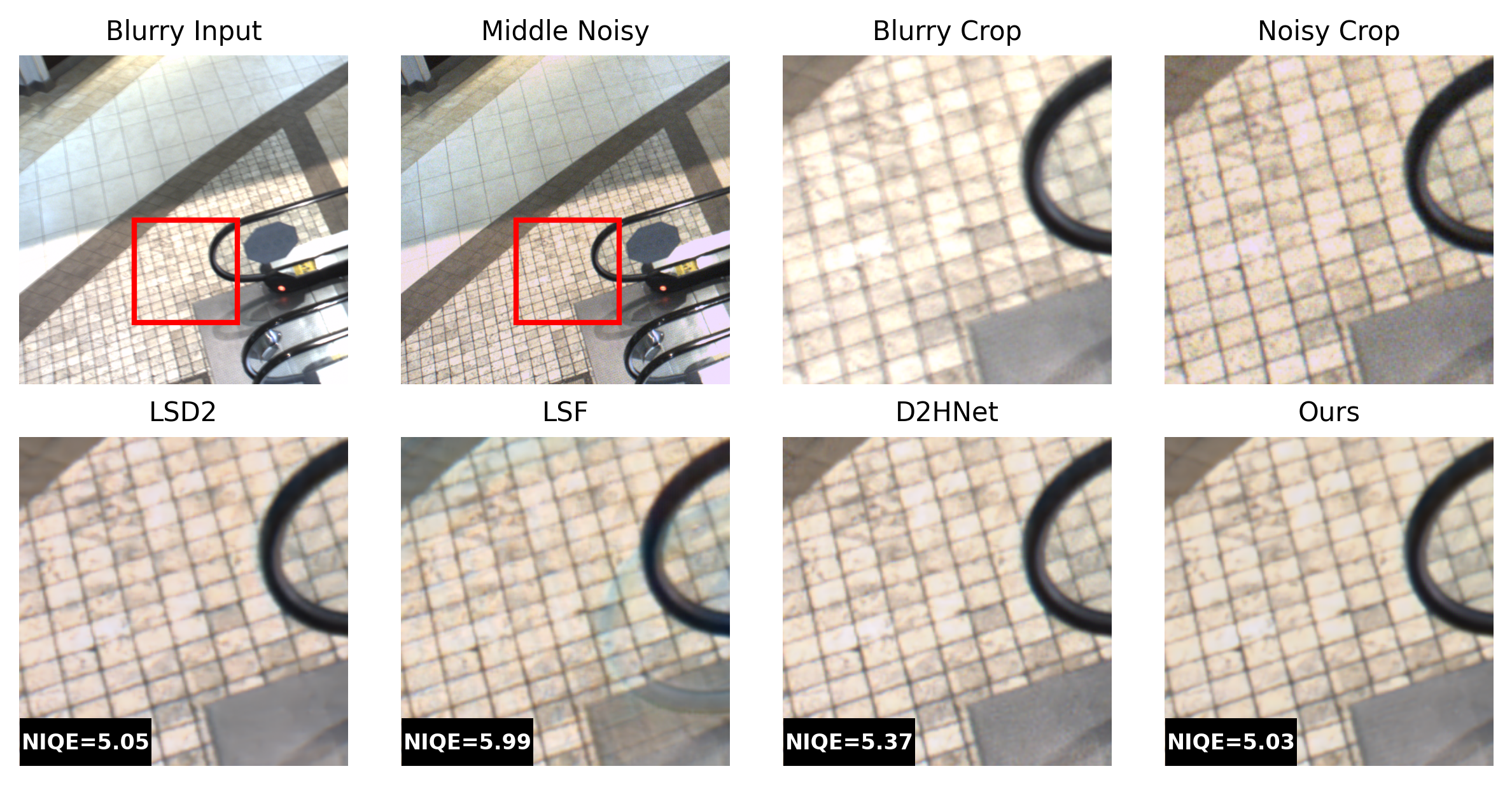}
     \end{subfigure}
    \begin{subfigure}[b]{\textwidth}
         \centering
         \includegraphics[width=\textwidth]{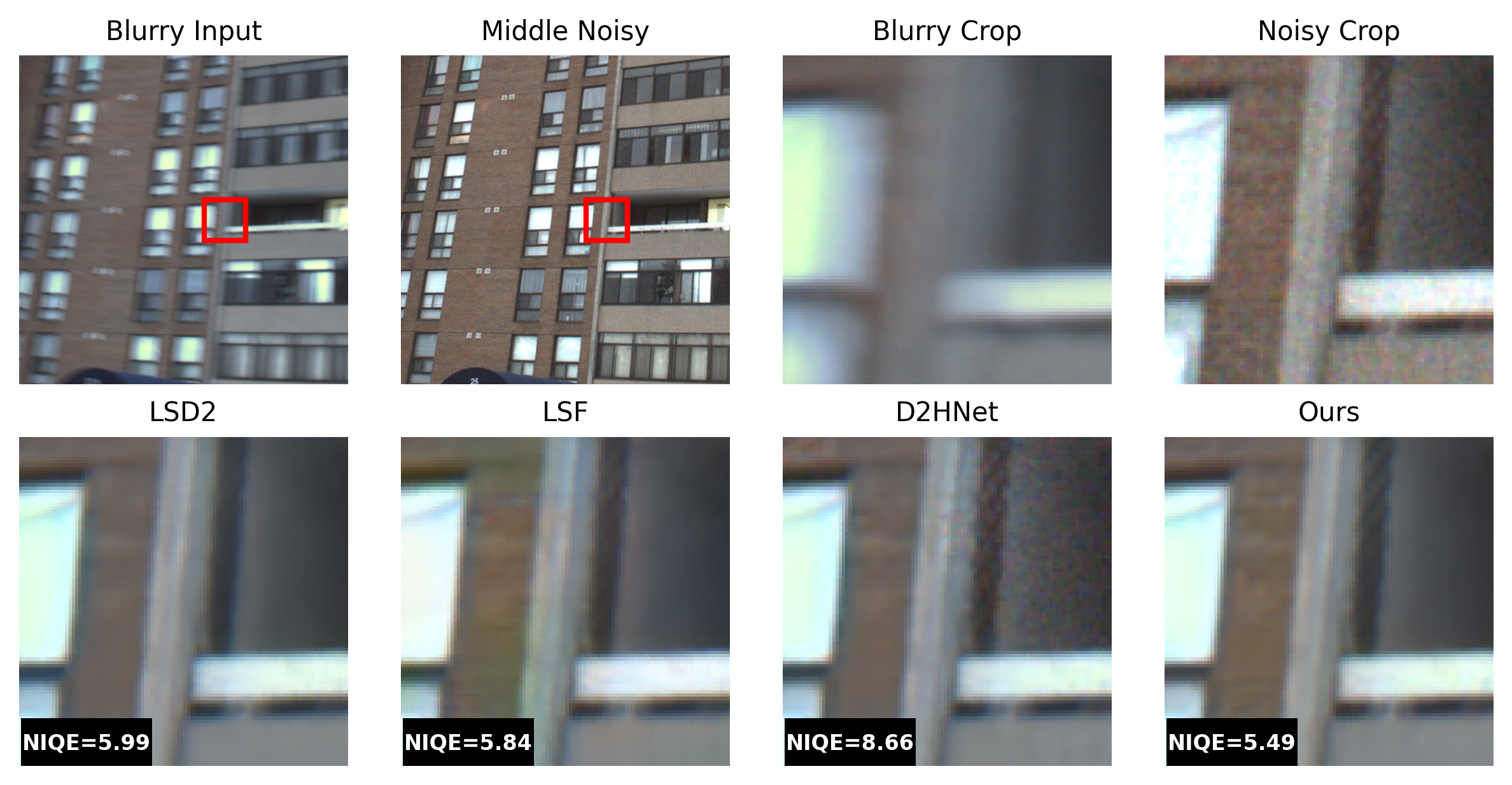}
     \end{subfigure}
     \caption{Qualitative comparison of our method with the state of the art on real data examples.}
     \vspace{-0.15cm}
     \label{fig:supp_real_joint}
\end{figure*}

In Fig.\ \ref{fig:supp_syn_joint} and Fig.\ \ref{fig:supp_real_joint}, we illustrate outputs of methods dedicated to the joint task of deblurring-denoising, on synthetic and real data, respectively. 
For synthetic data, the PSNR metric is reported showing that we outperform LSD2, LSF, and D2HNet for the provided examples. 
For real data, as the ground-truth is not available, we report Natural Image Quality Evaluator (NIQE), a non-reference-based metric commonly used for image quality assessment \cite{mittal2012making}.
A lower value of NIQE indicates better quality. We achieve the lowest value in all the provided examples.

\begin{figure*}
    \begin{subfigure}[b]{\textwidth}
         \centering
         \includegraphics[width=\textwidth]{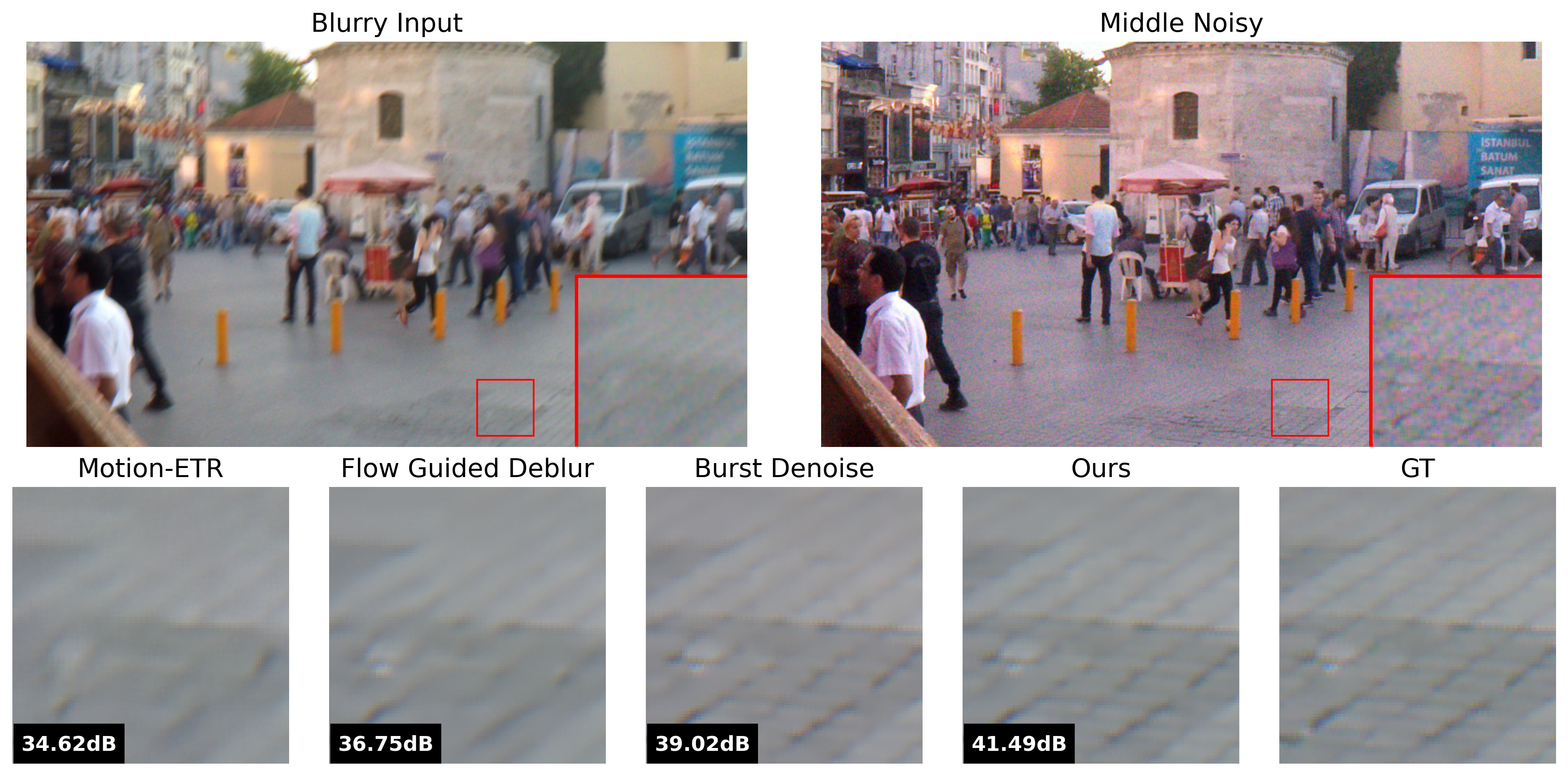}
     \end{subfigure}
    \begin{subfigure}[b]{\textwidth}
         \centering
         \includegraphics[width=\textwidth]{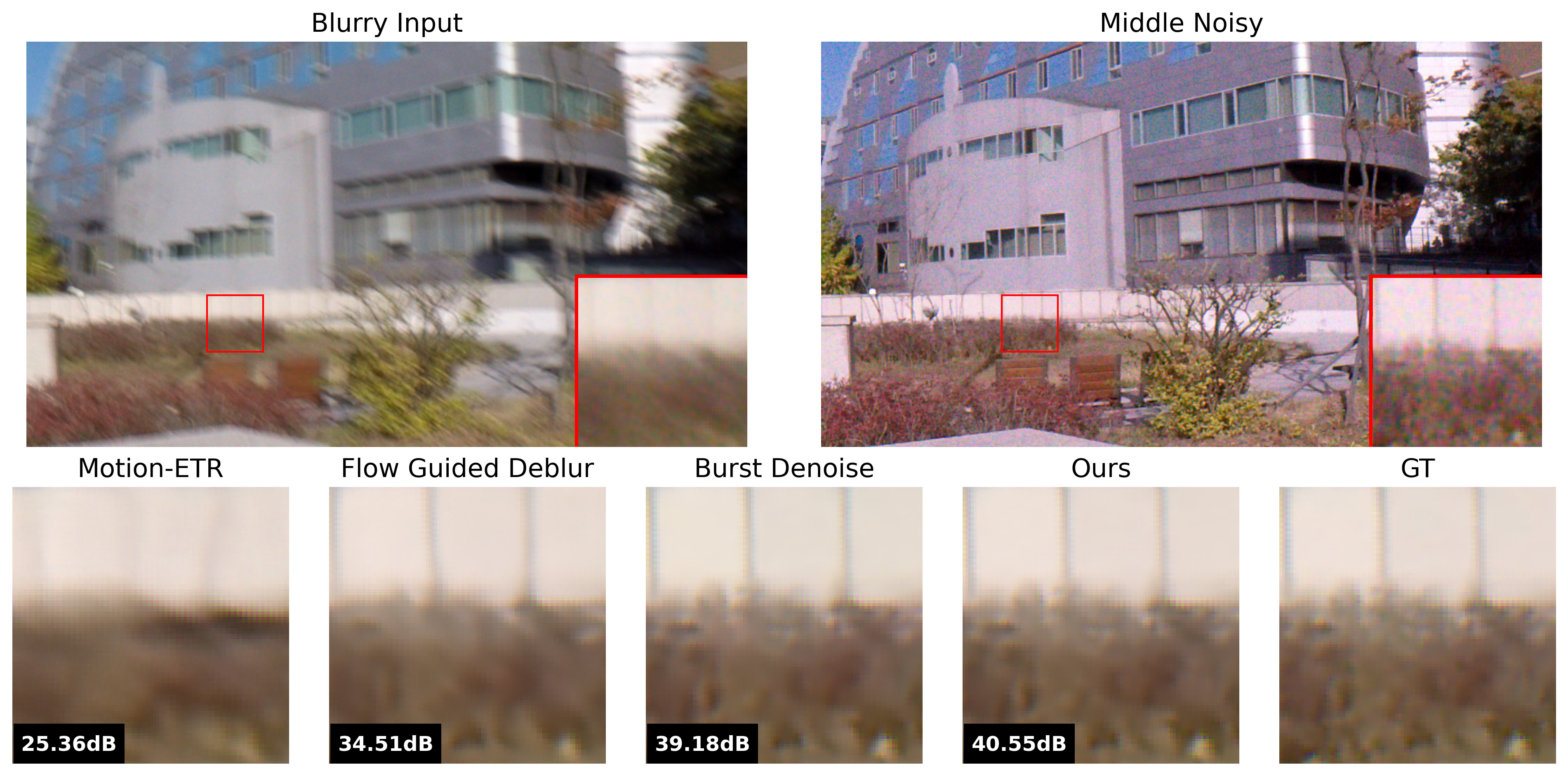}
     \end{subfigure}
     \caption{Qualitative comparison of our method with baselines that approach the individual tasks on synthetic data examples. Motion-ETR and Flow Guided Deblur correspond to deblurring of the input long exposure image, while Burst Denoise is the output of the denoiser trained separately on the burst images. Ours is using both the burst and long exposure images.}
     \vspace{-0.15cm}
     \label{fig:supp_syn_baselines}
\end{figure*}

\begin{figure*}
    \begin{subfigure}[b]{\textwidth}
         \centering
         \includegraphics[width=\textwidth]{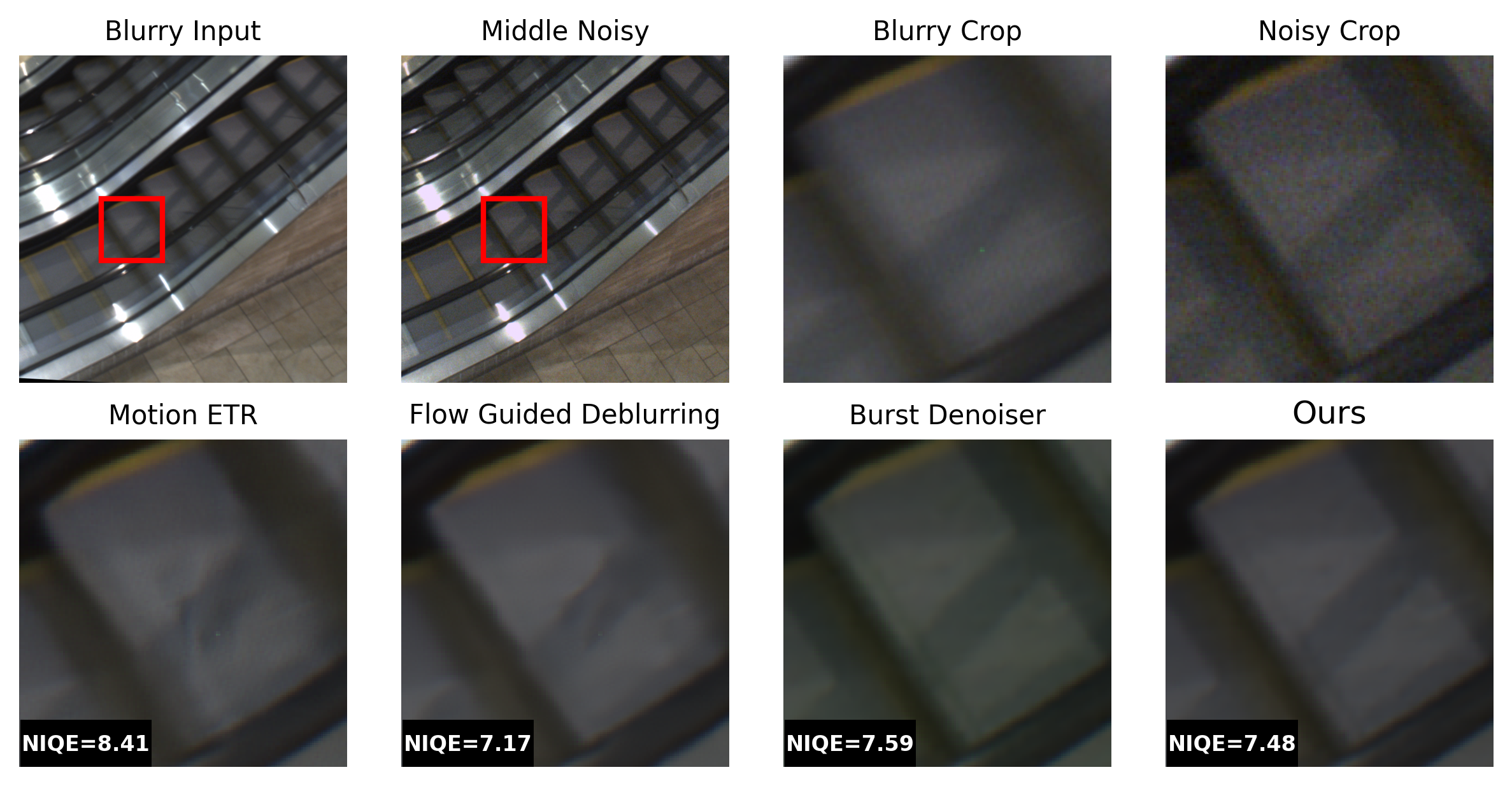}
     \end{subfigure}
    \begin{subfigure}[b]{\textwidth}
         \centering
         \includegraphics[width=\textwidth]{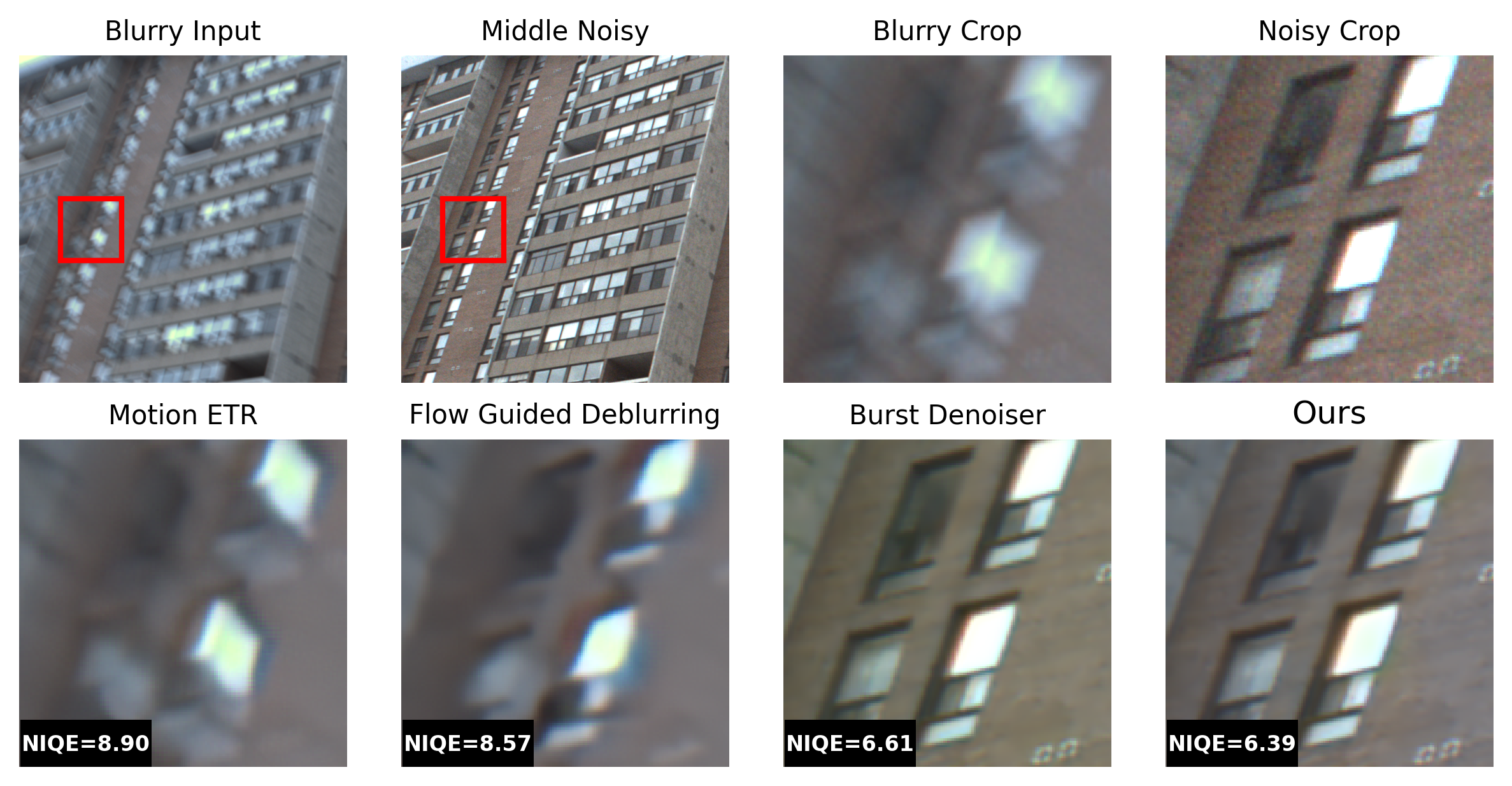}
     \end{subfigure}
     \caption{Qualitative comparison of our method with baselines that approach the individual tasks on real data examples. Motion-ETR and Flow Guided Deblur correspond to deblurring of the input long exposure image, while Burst Denoise is the output of the denoiser trained separately on the burst images. Ours is using both the burst and long exposure images.}
     \vspace{-0.15cm}
     \label{fig:supp_real_baselines}
\end{figure*}

Similarly, in Fig. \ref{fig:supp_syn_baselines} and Fig.\ \ref{fig:supp_real_baselines}, we compare our method with baselines that address the individual tasks of deblurring or denoising. PSNR and NIQE are used as quantitative metrics to evaluate all methods, on synthetic and real data, respectively.
In this comparison, our Flow-Guided Deblurring method outperforms Motion-ETR on both real and synthetic data. This validates that the optical flow is superior to the offsets provided by Motion-ETR.
Although the Burst Denoiser output is sharper than the deblurring outputs, it has noticeably wrong color information. 
Our method is able to mitigate this issue by extracting color information from the long exposure input.

\section{Network Architecture}
\label{sec:arch}
\subsection{Burst Denoiser}
Here, we describe the architecture of the burst denoiser module which is primarily based on previous work \cite{bhat2021deep}, in more detail.
Recall that our burst denoiser is composed of four sub-modules: encoder, alignment, fusion and decoder.
\vspace{0.15cm}

\textbf{Encoder.}
The encoder is shared between all frames in the burst. Given an RGB short exposure image, $S_i \in \mathbb{R}^{H \times W \times 3}$, the encoder starts with a conv layer ($3 \rightarrow 64$), followed by $9$ residual blocks ($64 \rightarrow 64)$. Then a single conv layer maps the output of residual blocks to a feature embedding with dimension $D=96$.
The final embedded burst features are denoted as $\{e_i\}_{i=1}^{N}, e_i \in \mathbb{R}^{H \times W \times 96}$.
\vspace{0.15cm}

\textbf{Alignment.}
The features, $e_i$, are not spatially aligned due to camera and scene motion.
We warp them back to the reference frame based on the optical flow between original images, $\Delta \mathbf{p_i} \in \mathbb{R}^{H \times W \times 2}$. Bilinear interpolation is used to backwarp the encoded features into aligned ones, $\{\tilde{e}_i\}_{i=1}^{N}, \tilde{e}_i \in \mathbb{R}^{H \times W \times 96}$.
\vspace{0.15cm}

\textbf{Fusion.}
Once aligned, the fusion sub-module combines the input features, $\tilde{e}_i$ through an attention mechanism, yielding a single feature map $e$.
In this sub-module, first, input features, $\tilde{e}_i$, are projected to a lower dimension, $D=64$, using a single conv layer.
From these projected features, denoted as $\tilde{e}_i^p \in \mathbb{R}^{H \times W \times 64}$, we compute residuals from the reference frame, $r_i = \tilde{e}_i^p - \tilde{e}_m^p$. 
As previous work \cite{bhat2021deep}, the residuals, $r_i$, are informative about alignment errors and thus let attention weights to be aware of misalignments.
We additionally extract features, denoted as $\Delta \mathbf{f_i} \in \mathbb{R}^{H \times W \ \times 64}$,
from the optical flow, $\Delta \mathbf{p_i}$, through a residual block. 
These features can provide information about sub-pixel sampling positions to the attention module.
We employ a residual network as the weight predictor that receives as input the reference features, $\tilde{e}_m^p$, plus the corresponding residual, $r_i$, and flow features, denoted as $\Delta \mathbf{f_i}$.
This network outputs unnormalized log weights as
\begin{equation}
    \tilde{w}_i=W(\tilde{e}_m^p, r_i, \Delta \mathbf{f_i}), \quad \tilde{w}_i \in \mathbb{R}^{H \times W} \ .
\end{equation}
To compute the final weights, the softmax function is applied in an element-wise fashion and then the weighted sum is computed across the burst.
This is summarized as follows:
\begin{equation}
    e = \sum_{i=1}^{N} w_i.\tilde{e}_i, \quad w_i = \frac{\exp(\tilde{w}_i)}{\sum_{j=1}^{N}\exp(\tilde{w}_j)} \ ,
\end{equation}
where $e \in \mathbb{R}^{H \times W \times 96}$ denotes the final feature.
\vspace{0.15cm}

\textbf{Decoder.}
The decoder maps the final merged features, $e$, into a final clean version of the reference image.
Our decoder first uses a conv layer to reduce the dimension of the fused features, $e$, to $D=64$, followed by a set of residual blocks. 
Then another conv layer lowers the dimension to $D=32$ and the resulting map is processed by a second set of residual blocks.
Eventually, a final conv layer predicts the final denoised RGB image. 
We also remove upsampling layers from the original decoder.

\subsection{Fusion in the Joint Task}

To meaningfully combine the information from deblurring and denoising networks, we adopted a fusion-based technique to generate the final output. The features obtained from deblur and denoise networks are fused using our fusion network to generate the final image. Our fusion network is motivated from from \cite{bhat2021deep}, uses a sequence of residual layers to merge the features into an enhanced RGB image. We have provided network details in Table.  \ref{table:fusion_arch} for reference.   

\begin{table}
\setlength{\tabcolsep}{1pt}
\resizebox{0.45\textwidth}{!}{%
\begin{tabular}{clcccl}
\toprule
\multicolumn{6}{c}{\textsc{Final Decoder}}                                                                                                                                                                                                           \\ \hline
\multicolumn{1}{l|}{\textbf{Layer \#}}           & \multicolumn{3}{c|}{\textbf{Layer description}}                                                                                                          & \multicolumn{2}{l}{\textbf{Output Shape}}                \\ \hline
\multicolumn{1}{c|}{0}                  & \multicolumn{3}{c|}{Input}                                                                                                                      & \multicolumn{2}{c}{$192 \times H \times W$ }                 \\ \hline
\multicolumn{1}{c|}{\multirow{2}{*}{1}} & \multicolumn{1}{l|}{\multirow{2}{*}{Pre-Decoder}}     & \multicolumn{2}{c|}{Conv2D $3 \times 3$ }                                                       & \multicolumn{2}{c}{\multirow{2}{*}{$96 \times H \times W$ }} \\ \cline{3-4}
\multicolumn{1}{c|}{}                   & \multicolumn{1}{l|}{}                                 & \multicolumn{2}{c|}{ReLU}                                                               & \multicolumn{2}{c}{}                            \\ \hline
\multicolumn{1}{c|}{\multirow{2}{*}{2}} & \multicolumn{1}{l|}{\multirow{2}{*}{Initial Layer}}     & \multicolumn{2}{c|}{Conv2D $3 \times 3$ }                                                       & \multicolumn{2}{c}{\multirow{2}{*}{$64 \times H \times W$ }} \\ \cline{3-4}
\multicolumn{1}{c|}{}                   & \multicolumn{1}{l|}{}                                 & \multicolumn{2}{c|}{ReLU}                                                               & \multicolumn{2}{c}{}                            \\ \hline
\multicolumn{1}{c|}{\multirow{8}{*}{3}} & \multicolumn{1}{l|}{\multirow{8}{*}{Pre-Residual Layer}}  & \multicolumn{1}{l|}{\multirow{4}{*}{Res. Block \#1}} & \multicolumn{1}{c|}{Conv2D $3 \times 3$ } & \multicolumn{2}{c}{\multirow{8}{*}{$64 \times H \times W$ }} \\ \cline{4-4}
\multicolumn{1}{c|}{}                   & \multicolumn{1}{l|}{}                                 & \multicolumn{1}{l|}{}                               & \multicolumn{1}{c|}{ReLU}         & \multicolumn{2}{c}{}                            \\ \cline{4-4}
\multicolumn{1}{c|}{}                   & \multicolumn{1}{l|}{}                                 & \multicolumn{1}{l|}{}                               & \multicolumn{1}{c|}{Conv2D $3 \times 3$ } & \multicolumn{2}{c}{}                            \\ \cline{4-4}
\multicolumn{1}{c|}{}                   & \multicolumn{1}{l|}{}                                 & \multicolumn{1}{l|}{}                               & \multicolumn{1}{c|}{ReLU}         & \multicolumn{2}{c}{}                            \\ \cline{3-4}
\multicolumn{1}{c|}{}                   & \multicolumn{1}{l|}{}                                 & \multicolumn{2}{c|}{Res. Block \#2}                                                      & \multicolumn{2}{c}{}                            \\ \cline{3-4}
\multicolumn{1}{c|}{}                   & \multicolumn{1}{l|}{}                                 & \multicolumn{2}{c|}{Res. Block \#3}                                                     & \multicolumn{2}{c}{}                            \\ \cline{3-4}
\multicolumn{1}{c|}{}                   & \multicolumn{1}{l|}{}                                 & \multicolumn{2}{c|}{Res. Block \#4}                                                     & \multicolumn{2}{c}{}                            \\ \cline{3-4}
\multicolumn{1}{c|}{}                   & \multicolumn{1}{l|}{}                                 & \multicolumn{2}{c|}{Res. Block \#5}                                                     & \multicolumn{2}{c}{}                            \\ \hline
\multicolumn{1}{c|}{\multirow{2}{*}{4}} & \multicolumn{1}{l|}{\multirow{2}{*}{Mid Layer}}       & \multicolumn{2}{c|}{Conv2D $3 \times 3$ }                                                       & \multicolumn{2}{c}{\multirow{2}{*}{$32 \times H \times W$ }} \\ \cline{3-4}
\multicolumn{1}{c|}{}                   & \multicolumn{1}{l|}{}                                 & \multicolumn{2}{c|}{ReLU}                                                               & \multicolumn{2}{c}{}                            \\ \hline
\multicolumn{1}{c|}{\multirow{5}{*}{5}} & \multicolumn{1}{l|}{\multirow{5}{*}{Post-Residual Layer}} & \multicolumn{2}{c|}{Res. Block \#1}                                                     & \multicolumn{2}{c}{\multirow{5}{*}{$32 \times H \times W$ }} \\ \cline{3-4}
\multicolumn{1}{c|}{}                   & \multicolumn{1}{l|}{}                                 & \multicolumn{2}{c|}{Res. Block \#2}                                                     & \multicolumn{2}{c}{}                            \\ \cline{3-4}
\multicolumn{1}{c|}{}                   & \multicolumn{1}{l|}{}                                 & \multicolumn{2}{c|}{Res. Block \#3}                                                     & \multicolumn{2}{c}{}                            \\ \cline{3-4}
\multicolumn{1}{c|}{}                   & \multicolumn{1}{l|}{}                                 & \multicolumn{2}{c|}{Res. Block \#4}                                                     & \multicolumn{2}{c}{}                            \\ \cline{3-4}
\multicolumn{1}{c|}{}                   & \multicolumn{1}{l|}{}                                 & \multicolumn{2}{c|}{Res. Block \#5}                                                     & \multicolumn{2}{c}{}                            \\ \hline
\multicolumn{1}{c|}{\multirow{2}{*}{6}} & \multicolumn{1}{l|}{\multirow{2}{*}{Predictor}}       & \multicolumn{2}{c|}{Conv2D $3 \times 3$ }                                                       & \multicolumn{2}{c}{\multirow{2}{*}{$3 \times H \times W$ }}  \\ \cline{3-4}
\multicolumn{1}{c|}{}                   & \multicolumn{1}{l|}{}                                 & \multicolumn{2}{c|}{ReLU}                                                               & \multicolumn{2}{c}{}                           
\end{tabular}%
}
\caption{Fusion Network Architecture. Our fusion network consists of residual blocks to process the input features from deblur and denoise networks and generate high quality image.}
\vspace{-0.5cm}
\label{table:fusion_arch}
\end{table}

\end{document}